\documentclass{article}

\usepackage[preprint, nonatbib]{nips_2018}

\usepackage{float}

\usepackage[square,numbers]{natbib}
\usepackage[utf8]{inputenc} 
\usepackage[T1]{fontenc}    
\usepackage{hyperref}       
\usepackage{url}            
\usepackage{booktabs}       
\usepackage{amsfonts}       
\usepackage{nicefrac}       
\usepackage{microtype}      
\usepackage{graphicx}
\usepackage{tabularx}
\usepackage{hyperref}
\usepackage{subcaption}
\usepackage{algorithm}
\usepackage{algpseudocode}
\usepackage{placeins}
\usepackage{amsmath}
\usepackage{tabularx}
\usepackage{booktabs}
\usepackage{multirow}
\title{Vision-Based Deep Reinforcement Learning of UAV Autonomous Navigation Using Privileged Information}

\author{
  Junqiao Wang \\
  \texttt{23S004019@stu.hit.edu.cn} \\
\And
 Zhongliang Yu \\
  \texttt{zlyu@cqu.edu.cn} \\
\And
Dong Zhou \\
  \texttt{dongzhou@hit.edu.cn} \\
  \And
Jiaqi Shi \\
\texttt{22s104188@stu.hit.edu.cn} \\
\And
Runran Deng \\
\texttt{deng\_run\_ran@126.com} \\
}

\begin{document}
\maketitle

\begin{abstract}
The capability of UAVs for efficient autonomous navigation and obstacle avoidance in complex and unknown environments is critical for applications in agricultural irrigation, disaster relief and logistics. In this paper, we propose the DPRL (Distributed Privileged Reinforcement Learning) navigation algorithm, an end-to-end policy designed to address the challenge of high-speed autonomous UAV navigation under partially observable environmental conditions. Our approach combines deep reinforcement learning with privileged learning to overcome the impact of observation data corruption caused by partial observability. We leverage an asymmetric Actor-Critic architecture to provide the agent with privileged information during training, which enhances the model’s perceptual capabilities. Additionally, we present a multi-agent exploration strategy across diverse environments to accelerate experience collection, which in turn expedites model convergence. We conducted extensive simulations across various scenarios, benchmarking our DPRL algorithm against the state-of-the-art navigation algorithms. The results consistently demonstrate the superior performance of our algorithm in terms of flight efficiency, robustness and overall success rate.
\end{abstract}

\section{Introduction}
Drones, also known as unmanned aerial vehicles (UAVs), have rapidly evolved in recent years, playing a pivotal role in various industries due to their efficiency and versatility\cite{kalidas2023deep}. One of the key advantages of drones is their ability to efficiently operate in challenging or unpredictable environments without endangering human lives\cite{rs15133266}, making them well-suited for tasks such as monitoring large agricultural fields\cite{SU2023242}, navigating complex logistics routes\cite{drones7050323}, and conducting post-disaster search and relief operations\cite{drones7030191}. As drones continue to demonstrate their value across various applications, the ability to autonomously navigate in unknown complex environments becomes increasingly critical and has gradually gained significant attention.\cite{yin2024autonomous,10.1109/ICRA48891.2023.10160563,10.1109/TIE.2024.3363761,10.3390/rs16030471}

Current autonomous drone navigation systems rely on a variety of sensors, including LiDAR, cameras, radar, GPS, RTK, and inertial measurement units (IMUs), to perceive with their environment\cite{drones7020089}. Among these sensors, cameras offer significant advantages due to their lightweight, low power consumption, and cost-effectiveness\cite{fei2024deep}, yet they provide high-resolution information about the surrounding environment, such as color and texture, making them preferable to other sensors like LiDAR for small drones with limited payload capacity\cite{10.1016/j.eswa.2017.09.033}. 
In recent years, numerous UAV visual navigation algorithms based on onboard cameras have emerged, utilizing visual information for obstacle detection and avoidance\cite{10.1109/IVS.2016.7535370,10.1109/ACCESS.2019.2953954}, visual odometry\cite{duan2022stereo,teed2024deep}, Simultaneous Localization and Mapping (SLAM)\cite{teed2021droid,sumikura2019openvslam}, motion planning\cite{lu2023lpnet,9213901}, and flight control\cite{kaufmann2023champion,bhattacharya2024vision}.

In visual navigation, traditional algorithms, though widely used, often rely on hand-crafted features and predefined models, making them less adaptable to highly dynamic and complex environments, especially in the presence of noisy sensory data\cite{srivastava2022edge,f14020268}. These methods generally require precise tuning and struggle with unforeseen obstacles or changing conditions, leading to decreased efficiency and safety. However, deep reinforcement learning (DRL) offers significant advantages for visual navigation by enabling systems to learn navigation policies directly from raw visual data through trial and error\cite{lu2023lpnet, 9907788}. This kind of method reduces the need for extensive feature engineering and allows drones to autonomously adapt to uncertainties in complex environments, such as dynamic obstacles\cite{tong2021uav} and varying lighting conditions\cite{loquercio2021learning}, thereby enhancing robustness and flexibility. 

Despite the recent advances made by deep reinforcement learning-based visual navigation algorithms, they often overlook the impact of partial observability in the environment, which can severely affect the decision-making effectiveness of these models\cite{joshi2024sim} and significantly degrade their performance when transitioning to real-world environments. Additionally, existing deep reinforcement learning-based visual navigation methods often struggle with low efficiency in experience collection, particularly in complex environments where successful flights are challenging to achieve, leading to slow model convergence\cite{he2020deep}.

\begin{figure}[t]
		\centering
		\includegraphics[scale=0.35]{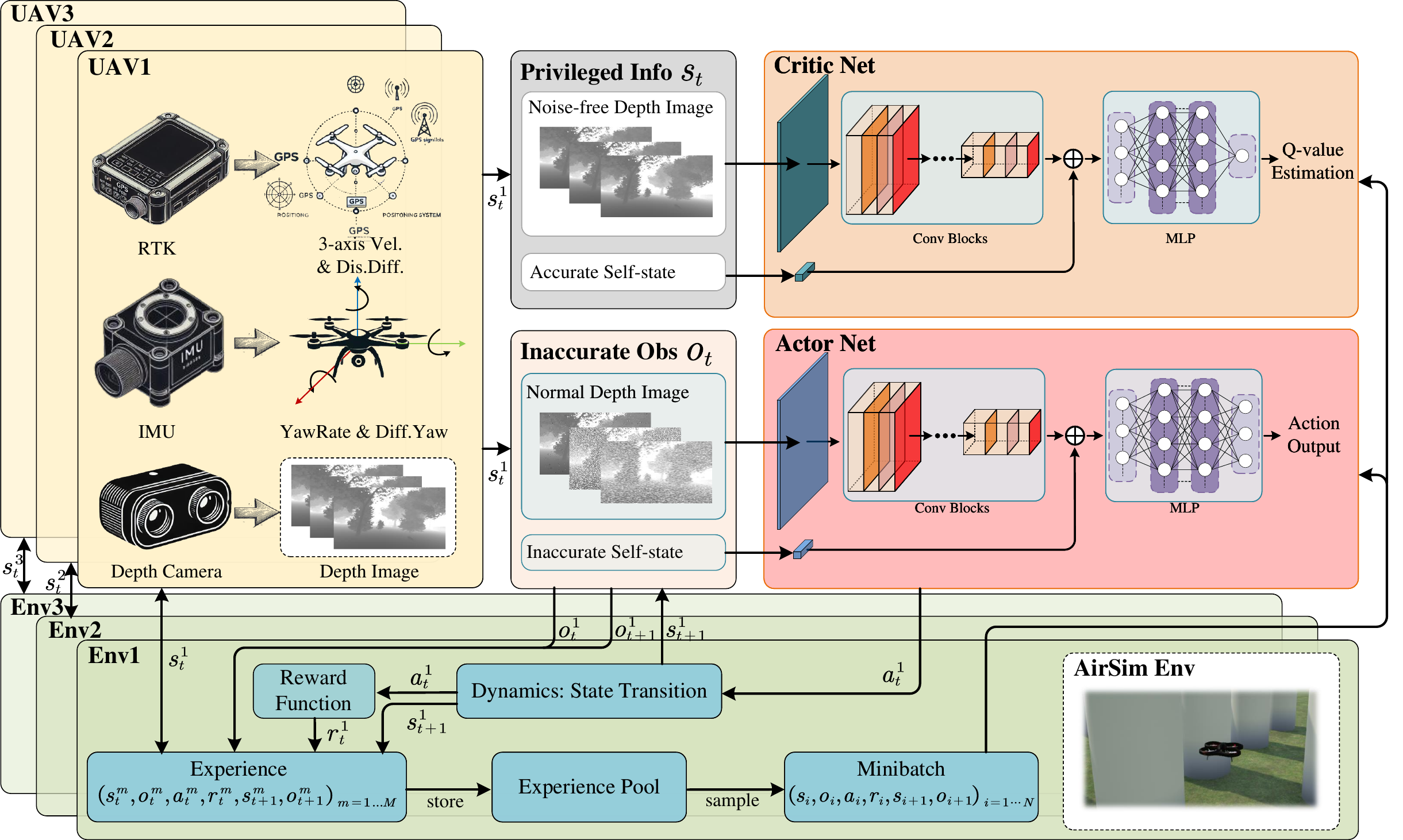}
	\caption{The DPRL Framework for UAV Navigation.\label{Framework}}
\end{figure}   

To address the above issues, we propose the DPRL algorithm for UAV visual navigation, the algorithm framework is shown in Figure~\ref{Framework}. DPRL incorporates an asymmetric Actor-Critic network structure, where the Critic network receives accurate, noise-free privileged perception information during training, enabling the model to build resilience against the interference in perception data caused by partial observability in the environment. Additionally, we propose a multi-agent exploration strategy which facilitates asynchronous experience collection across multiple environments to accelerate model convergence. This algorithm enables high-speed obstacle avoidance for UAVs in complex and unknown environments, while taking into account the inaccuracy of observations. The main contributions are summarized as follows:

\begin{enumerate}
	\item	We integrate deep reinforcement learning with privileged learning by providing accurate perception information to the Critic network during training, which greatly enhances the model's ability to handle environment uncertainties.
	\item	We propose a multi-agent exploration strategy where multiple UAVs asynchronously operate in simulated environments to collect experiences, enhancing efficiency and accelerating convergence.
	\item   We validated the algorithm's advantages in success rate, efficiency, and robustness over TD3 and EGO-Planner-v2, with ablation studies confirming its design effectiveness.
\end{enumerate}

The rest of the paper is organized as follows: Section 2 discusses related work on autonomous UAV navigation and obstacle avoidance algorithms. Section 3 introduces the proposed algorithm in this study. Section 4 describes the experimental setup, results, and analysis. Section 5 provides a conclusion and outlines future work.

\section{Related Work}
\subsection{Vision-based UAV Navigation}
Traditional approaches achieve autonomous UAV navigation and obstacle avoidance by performing sequential, cascaded tasks, often using visual navigation frameworks that leverage depth images to construct point cloud maps\cite{sumikura2019openvslam, teed2021droid,sarlin2020superglue}, which are then used for path planning\cite{li2022uav, guo2022fc, zhao2020path, ait2022novel, huang2022simulated} and flight control\cite{lindqvist2020nonlinear, mohammadi2023robust, 8864571, 9354996}. Fast-Planner developed by Shen et al.\cite{8758904,9196996,9422918}, follows this pattern, constructs a grid map by combining depth camera data with localization information, employs a hybrid A* algorithm for global path planning, optimizes trajectories using B-splines with automatic time allocation adjustments, and designs a controller for trajectory tracking. Recently, Zhou et al.\cite{zhou2022swarm} enhanced this framework and introduced EGO-Planner-v2, which achieves precise localization through visual-inertial odometry (VIO) based on grayscale images and IMU data, constructs a probabilistic map using depth images and localization data, conducts spatiotemporal trajectory planning based on this map and designs a controller to track the planned trajectory, enabling UAVs to navigate autonomously in complex outdoor environments.

While the frameworks mentioned above offer high interpretability and precise navigation with accurate maps, their cascading task structure often fails to capture interactions between tasks, leading to error accumulation. Additionally, map building and storage require substantial resources, lowering frame rates and limiting responsiveness to dynamic environmental changes, which reduces obstacle avoidance efficiency\cite{loquercio2021learning}. Recent advancements in deep learning have led to end-to-end visual navigation frameworks based on supervised learning that directly map perception to control\cite{loquercio2018dronet,gandhi2017learning,bhattacharya2024vision}. These models, though less interpretable, eliminate cumulative error and the need for map storage, saving computational resources and improving adaptability to dynamic environments. 

While supervised approaches struggle with data labeling and often overfit, deep reinforcement learning methods require no prior knowledge, instead learning through environmental interaction to develop robust obstacle avoidance strategies. Kalidas et al.\cite{kalidas2023deep} considered dynamic obstacle avoidance and established three simulation environments using AirSim. In these simulated environments, they trained and compared the performance of three deep reinforcement learning algorithms: DQN, PPO, and SAC. Myoung et al.\cite{lee2021deep} incorporated the Hindsight Experience Replay (HER) algorithm into the SAC framework to address the issue where the maximum entropy optimization objective in SAC may reduce the optimality of policy. They compared the proposed SACHER algorithm with SAC and DDPG, demonstrating its superiority. He et al.\cite{he2021explainable} proposed a model explanation method based on feature attribution to analyze the mapping relationship between features and decisions in reinforcement learning policy networks during the learning process. This approach enables better adjustment of network structures to enhance model performance. Their algorithm was validated in both simulation and real-world environments.

\subsection{Strategies to Accelerate DRL Convergence}
One of the major challenges in deep reinforcement learning algorithms is the low efficiency of experience collection, which leads to slow model convergence. Researchers have proposed various methods to address this issue. To tackle the inefficiency of experience selection, Hu et al.\cite{10149507} combined curriculum learning with an improved prioritized experience replay (PER) method by launching independent threads to assign sampling priorities based on the current curriculum difficulty and the TD error of the experiences. They further eliminated low-priority experiences to improve the efficiency of selecting valuable experiences for model updates. However, curriculum learning is often limited by the rationality of difficulty settings, and significant changes in the distribution of observed data after curriculum transitions can cause noticeable performance degradation in the model.

Given the challenges in ensuring the effectiveness of curriculum learning's difficulty settings and transition mechanisms, many studies have shifted their focus toward learning from demonstration. He et al.\cite{he2020deep} adopted imitation learning by using the 3DVHF* algorithm as an expert policy. They pre-trained the Actor and Critic networks of the reinforcement learning model in a supervised manner using expert-generated experiences before allowing the model to interact with the environment to further enhance its navigation and obstacle avoidance capabilities. However, the interaction experiences significantly impact the pre-trained weights, requiring partial layers of the decision model to be frozen to prevent large parameter changes.

Due to the problem of model performance degradation caused by significant changes in training scenarios and experience distributions, many studies have turned to refining action selection methods. Xie et al.\cite{xie2021unmanned} modified the action selection strategy by building on the $\epsilon$-greedy method. They incorporated rewards and Q-value estimates to select actions, avoiding errors caused by inaccurate Q-value estimates during the early training stages. This ensures that UAVs maximize their movement in the early training phase to collect diverse environmental data. However, this strategy is only suitable for discrete action spaces. In cases with large or continuous action spaces, it becomes difficult to effectively evaluate all possible actions, leading to reduced efficiency.

To efficiently utilize incomplete and noisy perceptual information while ensuring accelerated model convergence and robust decision-making, many studies have explored the application of privileged learning. Kaufmann et al.\cite{kaufmann2023champion} addressed the issue of estimation errors in VIO by employing privileged learning, where the Critic network is provided with accurate pose and velocity information during model training. This approach enables the effective use of inaccurate experiences, thereby accelerating convergence. However, in UAV visual navigation tasks where environmental information is only partially observable, providing accurate pose data alone is insufficient to mitigate the impact of perception noise. To address this, we employed an asymmetric Actor-Critic structure, supplying the Critic network with all accurate observational data, including noise-free depth images, as privileged information to counteract the effects of corrupted observations due to partial observability. Furthermore, we proposed a multi-agent exploration strategy in which multiple agents asynchronously collect experiences across various environments, populating a centralized experience replay buffer. This buffer is used to update a central model, significantly improving the efficiency of experience collection and accelerating model convergence.
\section{Methodology}
The framework of our proposed DPRL navigation algorithm is illustrated in Figure~\ref{Framework}. In this framework, each UAV independently and asynchronously collects interaction experiences within its respective environment. Accurate state data are obtained through onboard sensors and subsequently perturbed with noise to simulate observations under partially observable environmental conditions. The accurate and inaccurate observation data are used as privileged information and normal sensory input, respectively, and are fed into the Critic and Actor networks. The Critic network evaluates the state-action pair's value, while the Actor network generates action outputs based on the current observations to control each UAV's flight in its designated environment.

\subsection{Problem Formulation}
In this section, we first model the autonomous navigation task of an unmanned aerial vehicle (UAV) as a Partially Observable Markov Decision Process (POMDP), following the modeling approach of Zhou et al. \cite{10143684}. Building on this framework, we then discuss the key components of reinforcement learning (RL) that drive the UAV's autonomous navigation.

A Partially Observable Markov Decision Process is characterized by a tuple $< s,o,a,r,p,\gamma>$ , representing states, observations, actions, transition probabilities, rewards, and discount factor, respectively. For the UAV navigation task, we define these elements as follows:
\begin{enumerate}
	\item \textbf{State}($s$): The state represents the current environment context and UAV status, which may include the UAV's position, velocity, orientation, proximity to obstacles, and sensor readings. 
	\item \textbf{Observation}($o$):The Observation refers to the information gathered by an agent to infer the state of its environment. In partially observable environments, the UAV cannot directly access the full state information. Instead, it depends on observations collected from its onboard sensors, which offer a noisy and incomplete representation of the surrounding environment.
	\item \textbf{Action}($a$): The action space consists of the possible control commands that the UAV can take at each time step. These actions may include altering the UAV's speed, direction, or altitude.
	\item \textbf{Transition Probability}($p$): The transition probability defines the likelihood of reaching a new state given the current state and an action. In the context of UAV autonomous navigation, the state transition is determined by the UAV's dynamics, which is influenced by factors such as control inputs, environmental conditions, and system noise.
	\item \textbf{Reward}($r$): The reward function evaluates the immediate performance of the UAV based on its action in a particular state. The objective is to design a reward function that encourages the UAV to navigate efficiently towards its destination while avoiding collisions. 
	\item \textbf{Discount Factor}($\gamma$): The discount factor determines the importance of future rewards relative to immediate rewards. A value close to 1 implies that future rewards are highly significant, which helps in encouraging long-term goal achievement.
\end{enumerate}

The key components of reinforcement learning are the agent, policy, and environment. The agent refers to the entity that performs actions and interacts with the environment; in the context of UAV autonomous navigation, this is the UAV itself. The policy serves as the decision-making guideline, which enables the agent to select actions based on its observations of the environment. The environment encompasses all external factors that the UAV must navigate, including terrain, obstacles, and dynamic elements that influence the UAV's state.

At each time step, the agent selects an action from a predefined action space, receives feedback in the form of a reward, and transitions to a new state based on the environment's response. This cycle continues as the agent gathers experience to refine its policy, ultimately learning an optimal strategy for safe and efficient navigation in complex environments.

\subsection{Privileged Reinforcement Learning}
In this section, we present the core of the UAV autonomous navigation framework: the deep reinforcement learning policy network integrated with privileged learning. We begin by introducing the reward function design tailored to this study, which plays a critical role in guiding the UAV's behavior. Next, we detail the integration of deep reinforcement learning with privileged learning, explaining how privileged information is utilized during training to enhance model performance, along with the policy update mechanism. Finally, we describe the specific network architecture designed to effectively implement the proposed framework and handle the challenges of partially observable environments.

\subsubsection{Reward Function Design}
In our work, We have designed a reward function that incorporates both continuous and sparse rewards. At the end of each training episode, the agent receives sparse rewards, including a goal-reaching reward, a collision penalty, and an out-of-bounds penalty. Specifically, when the drone enters the target zone (within 2 meters), it is awarded a positive reward of $+10$. Conversely, if the drone collides with obstacles or exceeds the environment boundaries, it incurs a negative reward of $-5$.

In addition, the agent is given continuous rewards at each time step. These consist of a distance differential reward, a distance error penalty, and a collision proximity penalty. The distance differential reward evaluates the reduction in distance between the agent and the target from the previous time step to the current one. The distance error penalty is computed based on the distance between the drone and the straight line connecting the starting point and the target at each time step. The collision proximity penalty is determined by the difference between the shortest distance from the drone to any obstacle and the predefined safety distance at each time step. The expression for the continuous reward is shown in equation \ref{eq1}.

\begin{equation}
	\begin{cases}
		r_{e} = \frac{d_{t-1} - d_t}{d_{g}} \\
		p_{p} =  \left| \mathrm{clip}\left( \frac{d_{l}}{10}, 0, 1 \right) \right| + 
		2 \cdot \left| \mathrm{clip}\left( \frac{z - z_g}{5}, -1, 1 \right) \right| \\
		p_{o} = \left\{
		\begin{aligned}
			& 1 - \mathrm{clip}\left( \frac{d_{o} - d_{c}}{d_{s} - d_{c}}, 0, 1 \right) 
			&& \text{if }\, d_{o} < d_{s} \\
			& 0 
			&& \text{if } \, d_{o} \geq d_{s}
		\end{aligned}
		\right. \\
		r = \mathrm{clip}\left( \eta_r \cdot r_{e} - \eta_p \cdot p_{p} - \eta_o \cdot p_{o}, -1, 1 \right)
	\end{cases}
	\label{eq1}
\end{equation}

The term $d_{g}$ represents the distance from the starting point to the target, while $d_t$ and $d_{t-1}$ refer to the drone's current and previous distances from the target, respectively. The variables $z$ and $z_g$ represent the current and target coordinates of the drone along the z-axis, respectively, and $d_{l}$ denotes the distance between the drone's current position and the straight line connecting the starting point and the target. The function $\mathrm{clip}$ is a limiting function. $d_{o}$ represents the drone's closest distance to the surface of an obstacle, calculated from depth images when privileged learning is not applied. However, when privileged learning is applied, it is determined using the global obstacle map and the UAV's current position, ensuring greater accuracy. $d_{c}$ is the collision distance, and a collision is considered to occur if $d_{o}$ is less than $d_{c}$. $d_{s}$ is the safety distance, and the drone is considered at risk of collision if $d_{o}$ is smaller than $d_{s}$. $\eta_r$, $\eta_p$, and $\eta_o$ are the scaling factors for the reward and the two penalty terms, respectively.

The design of this continuous reward function encourages the drone to navigate towards the target through positive reward terms, while the distance error penalty ensures that the drone follows the shortest straight-line path and improves the accuracy of final navigation. The collision proximity penalty helps prevent the drone from getting too close to obstacles, which is particularly important in dense obstacle environments. Normalizing the continuous reward can help avoid large biases and, at the same time, highlight the weight of the sparse reward, making it more effective in providing global guidance and balancing exploration with exploitation.

\subsubsection{Policy Learning with Privileged Information}
Privileged Learning\cite{vapnik2009new} is a machine learning paradigm inspired by a common phenomenon in human learning: when learning new tasks, humans often benefit from additional information such as explanations, books, or demonstrations—resources that may not always be available during actual task execution. In machine learning, this additional information is known as "privileged information." Privileged Learning is often implemented using the Learning Using Privileged Information (LUPI) framework, where extra information is used during training to guide the model's learning, resulting in improved performance even when this information is not available at test time. 

In our work, to address the issue of inaccurate perception caused by partial observability, we provide the policy model with the following two types of privileged information during the training process:

\begin{enumerate}
	\item \textbf{Accurate Perception Information}: This includes providing the agent with unnoised depth images, localization information, and self-state perception data.
	\item \textbf{Obstacle Map}: This means providing the agent with prior knowledge of the global positions of obstacles. Using this map information, the agent can calculate the UAV's current closest distance to the obstacle surface to determine the collision proximity penalty in the reward function, thereby enabling more accurate obstacle avoidance guidance.
\end{enumerate}

We constructed an asymmetric Actor-Critic network structure, where the Critic Network receives accurate state information and the Actor Network receives partially observable information, as shown in Figure~\ref{Framework}. This approach allows the agent to gain accurate perception of the environment during training, leading to more precise value estimation of the action-state pairs. As a result, this helps accelerate model convergence and improves the success rate.

\begin{figure}[H]
	\centering
	\begin{minipage}[b]{0.47\linewidth}
		\centering
		\includegraphics[width=\linewidth]{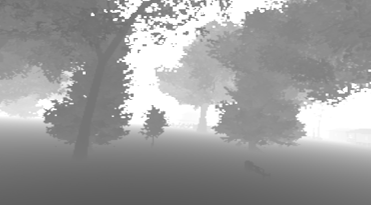}
		\caption*{(\textbf{a})}
	\end{minipage}
	\hfill
	\begin{minipage}[b]{0.47\linewidth}
		\centering
		\includegraphics[width=\linewidth]{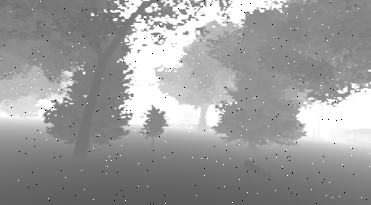}
		\caption*{(\textbf{b})}
	\end{minipage}
	\vskip 0.4cm
	\begin{minipage}[b]{0.47\linewidth}
		\centering
		\includegraphics[width=\linewidth]{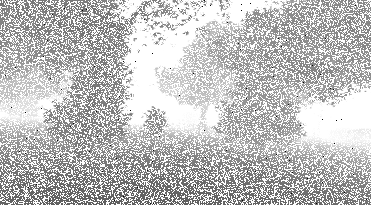}
		\caption*{(\textbf{c})}
	\end{minipage}
	\hfill
	\begin{minipage}[b]{0.47\linewidth}
		\centering
		\includegraphics[width=\linewidth]{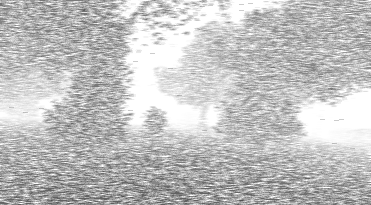}
		\caption*{(\textbf{d})}
	\end{minipage}
	\caption{The effect of adding noise to visual perception data. (\textbf{a}) Depth image obtained from the camera in the simulation environment. (\textbf{b}) Salt-and-pepper noise added to the depth image. (\textbf{c}) Gaussian noise added to image (\textbf{b}). (\textbf{d}) Motion blur applied to image (\textbf{c}).\label{noise}}
\end{figure}

In the simulation environment, we simulate the partially observable phenomena commonly encountered in the real world using different types of noise. Specifically, Gaussian noise with a mean of $\mu_s$
and a standard deviation of $\sigma_s$ is added to each dimension of localization and other self-state information, simulating sensor inaccuracies or estimation errors from VIO algorithms. To ensure stability, the noise is clipped to limit excessive interference. Additionally, noise is sequentially applied to depth image perception data in the specified order below, and the impact of progressively adding these three types of noise is illustrated in Figure~\ref{noise}.

\begin{enumerate}
	\item \textbf{Salt-and-Pepper Noise}: First, we introduce salt-and-pepper noise with a probability $p_{sp}$, randomly selecting pixels and setting them to extreme values (either 0 or 255). This simulates sudden sensor errors, such as those caused by abrupt lighting changes, strong reflections, or signal loss, where certain pixels become overly bright or dark.
	\item \textbf{Gaussian Noise}: After adding salt-and-pepper noise, we apply Gaussian noise with a mean $\mu_g$ and a standard deviation $\sigma_g$, representing overall sensor error. This introduces random measurement deviations across the entire image. The Gaussian noise blends with the salt-and-pepper noise, softening some of the extreme values and simulating a more realistic scenario where multiple types of noise coexist.
	\item \textbf{Motion Blur}: Lastly, we apply motion blur using a convolution operation with a kernel size $k_{mb}$. This simulates the blurring effect caused by either camera or object movement. The blur is applied after the other noises to simulate how, in real-world scenarios, noise caused by sudden disturbances would be further exacerbated by motion blur due to the UAV's high-speed movement or onboard camera jitter.
\end{enumerate}

\begin{figure}[H]
	\centering
	\includegraphics[scale=0.6]{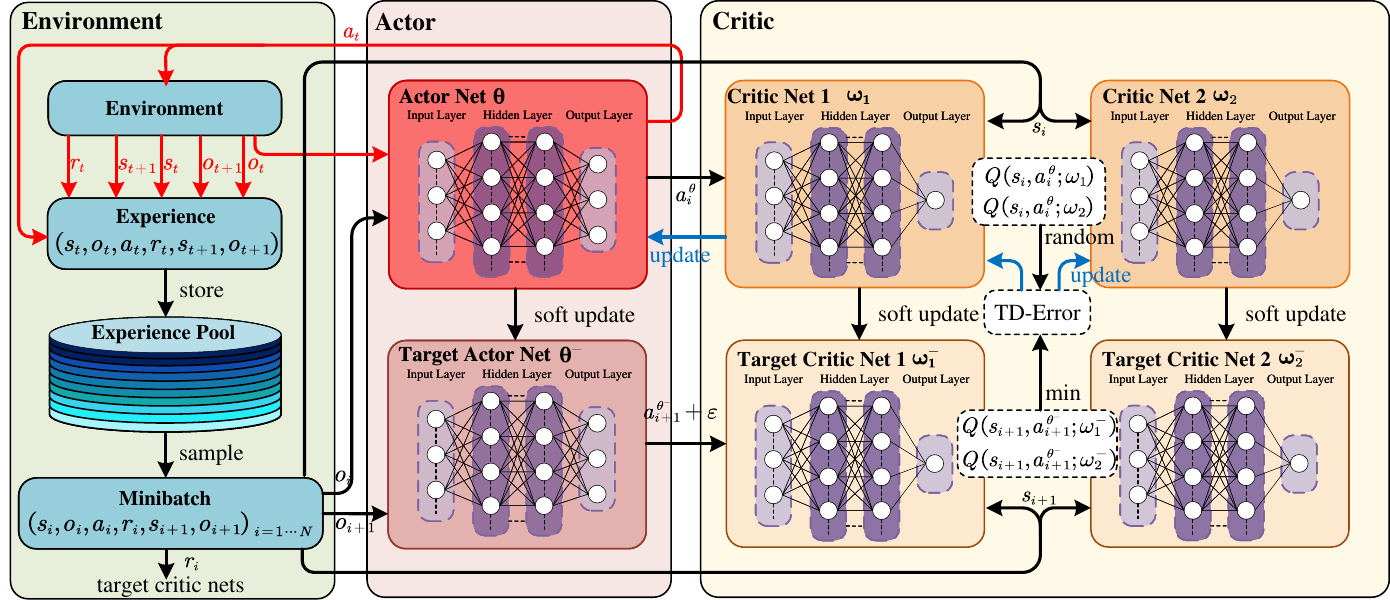}
	\caption{TD3 Framework within the POMDP Model.\label{td3}}
\end{figure}   

To ensure fast and stable model convergence, we perform network parameter updates based on the Twin Delayed Deep Deterministic Policy Gradient (TD3) algorithm\cite{fujimoto2018addressing}, which is an advanced version of the Deep Deterministic Policy Gradient (DDPG) algorithm\cite{lillicrap2015continuous}, specifically designed to address the instability issues inherent in DDPG. TD3 enhances the robustness of learning by implementing three significant improvements over DDPG, In the context of POMDPs, which include:

\begin{enumerate}
	\item \textbf{Clipped Double Q-learning}: TD3 consists of six networks: the Actor Network with parameters $\theta$, the Target Actor Network with parameter $\theta^-$, two Critic Networks with patameters $\omega$, and their corresponding Target Critic Networks with parameters $\omega^-$, as shown in Figure~\ref{td3}. At each step, $a_i^{\theta}$ represents the action selected by the Actor Network based on the current observation $o_i$, while $a_{i+1}^{\theta^-}$ is the action chosen by the Target Actor Network based on the next observation $o_{i+1}$. The accurate states corresponding to these observations, $s_i$ and $s_{i+1}$, align with $o_i$ and $o_{i+1}$ at their respective time steps. To address the overestimation bias present in DDPG, TD3 utilizes the two Critic Networks $Q(s_i,a_i^\theta;\omega_1)$ and $Q(s_i,a_i^\theta;\omega_2)$ to estimate the expected reward for a given state-action pair $(s_i, a_i)$. When calculating the TD target, the minimum value between the two Q-functions from the Target Critic Networks $Q(s_{i\operatorname{+}1},a_{i\operatorname{+}1}^{\theta^-};\omega_1^-)$ and $Q(s_{i\operatorname{+}1},a_{i\operatorname{+}1}^{\theta^-};\omega_2^-)$ is selected to avoid overestimation. The Critic Network loss function is given by:
	\begin{equation}
		L(\omega)=\mathbb{E}\left[\frac{1}{2}\left(R+\gamma\min_{j=1,2}Q\left(s_{i+1},a_{i+1}^{\theta-};\omega_j^-\right)-Q\left(s_i,a_i^{\theta};\omega\right)\right)^2\right]
		\label{critic_loss}
	\end{equation}
	Here, $R$ is the reward for the current step, $\gamma$ is the discount factor,  $\mathbb{E}[\cdot]$ denotes the expected TD error over a batch of replay data. 
	
	\item \textbf{Delayed Policy Updates}: TD3 employs a delayed update strategy for the Actor (Policy) Network. Unlike DDPG, where the policy and value networks are updated simultaneously, TD3 updates the Actor Network less frequently. This delay ensures that the Critic Network has had sufficient time to learn a more accurate Q-function before the policy is updated, preventing the Actor from being optimized using noisy or unstable value estimates. The policy loss function for the Actor Network is:
	\begin{equation}
		L(\theta) = -\frac1N\sum\left(\min_{j=1,2}Q(s,a^\theta;\omega_j)\right)
		\label{actor_loss}
	\end{equation}
	Here, $N$ is the batch size of replay data. This loss function maximizes the expected Q-value by selecting actions that result in higher rewards as predicted by the Critic Networks.
	
	\item \textbf{Target Policy Smoothing}: When calculating the TD target, noise is added to the output of the Target Actor Network $\pi(o_{i+1};\theta^-)$ to smooth the Q-values and prevent the policy from overfitting to sharp changes in the Q-function. The action from the Target Actor Network is perturbed by adding clipped noise $\varepsilon$, ensuring smoother policy updates:
	\begin{equation}
		\begin{aligned}
			&a_{i+1}' = \pi(o_{i+1};\theta^-) + \varepsilon \\
			&\varepsilon = \text{clip}(\mathcal{N}(0, \sigma), -c, c)
		\end{aligned}
	\end{equation}
	Here, $\varepsilon$ is sampled from a normal distribution $\mathcal{N}(0, \sigma)$ and clipped to the range $[-c, c]$.
\end{enumerate}

\begin{figure}[H]
	\centering
	\includegraphics[scale=0.65]{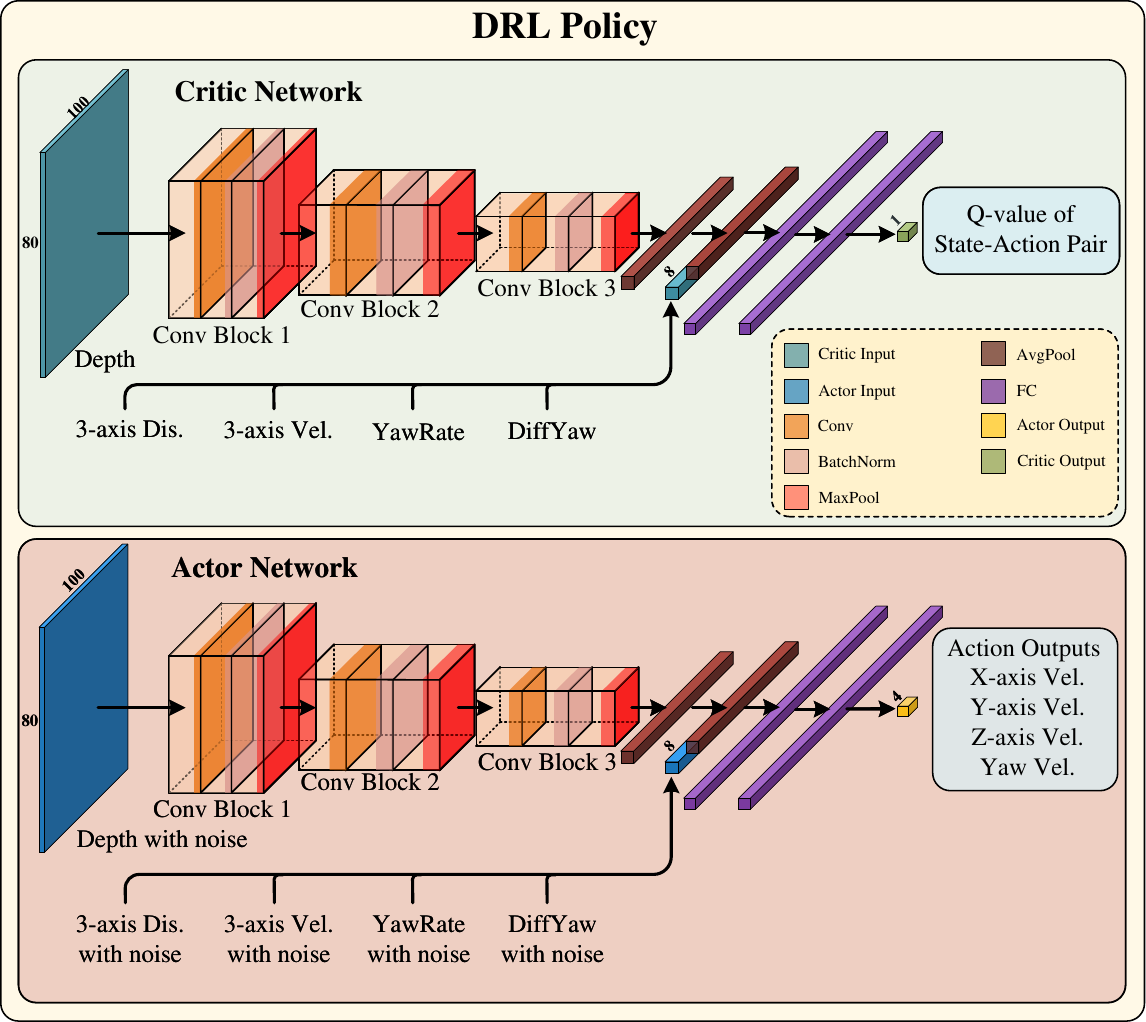}
	\caption{Architecture of Actor and Critic Network.\label{network}}
\end{figure}   

\subsubsection{Neural Network Architecture}
In our work, the network architecture is shown in Figure~\ref{network}. The state input utilizes multimodal information including visual perception information and UAV self-state variables. Specifically, the visual perception information is a depth image captured at a resolution of $240\times 320$ and resized to $80\times 100$ for storage and processing, followed by a feature extraction network that extracts features from the depth image and performs dimensionality reduction. The visual perception data $D\in\mathbb{R}^{80\times100}$ is compressed into $S_1\in\mathbb{R}^{25}$, which forms part of the overall system state. Another part is the 8-dimensional UAV self-state information vector $S_2\in\mathbb{R}^{8}$ consisting of the three-axis distances to the target point $[d_x, d_y, d_z] \in \mathbb{R}^3$, three-axis velocities$[v_x, v_y, v_z] \in \mathbb{R}^3$, the yaw angle deviation between the flight direction and the target direction $\Delta\psi\in\mathbb{R}^1$, and the current yaw angular velocity $\dot{\psi}\in \mathbb{R}^{1}$. From the above, the overall state vector is $S=[S_1, S_2]\in\mathbb{R}^{33}$.

It is worth noting that we use depth images as the visual perception data instead of RGB images. This decision is based on the fact that RGB images obtained in simulation environments often differ in color and texture from those in real-world settings. As a result, feature extraction networks trained in simulations may struggle to effectively extract features from real-world RGB images. In contrast, depth images provide only contour and distance information of obstacles within the field of view, which remains largely consistent between simulation and real environments. This consistency facilitates a smoother transition from simulation to reality and aids in obstacle avoidance through depth information.

On the other hand, the design of the 8-dimensional self-state vector aligns well with the data available from sensors in real-world flight scenarios. For instance, the three-axis position differences and three-axis velocity can be obtained via RTK modules or visual-inertial odometry (VIO) algorithms, while the yaw angle error and yaw rate can be measured using the IMU. Therefore, this state vector design is highly suitable for transitioning from simulation to real-world applications, ensuring both practicality and sufficiency for navigation and obstacle avoidance tasks without introducing unnecessary complexity.

As shown in Figure~\ref{network} and Table~\ref{net}, the policy network consists of two parts: a feature extraction network and a decision network. The feature extraction network is composed of CNN blocks that utilize CNN layers, BatchNorm and MaxPool, effectively extracting features from depth images, with ReLU serving as the activation function. The decision network is a two-layer MLP with 128 neurons per layer, which maps the state feature vectors to actions. Leaky ReLU is employed as the activation function to prevent issues with gradient vanishing that can occur with tanh activation in the TD3 algorithm, which may cause the Actor Network to continuously output boundary values.

The output of the network is a flight control command that belongs to the action space $A\in\mathbb{R}^4$, which consists of control commands for velocity $[v_x, v_y, v_z]\in\mathbb{R}^3$ and yaw rate $\dot{\psi}\in\mathbb{R}^1$. These continuous actions are used to directly control the UAV's motion, including adjusting speed and yaw rate. 

\begin{table}[t] 
	\caption{Network Architecture \label{net}}
	\newcolumntype{C}{>{\raggedright\arraybackslash}X}
	\begin{tabularx}{\textwidth}{CCCC}
		\toprule
		\textbf{Operator}     & \textbf{Input}       & \textbf{Filters}                                   & \textbf{Output}       \\
		\midrule
		Conv2D                & $1 \times 80 \times 100$ & $3 \times 3, 8$, stride 1                          & $8 \times 80 \times 100$   \\
		Max Pooling           & $8 \times 80 \times 100$ & $2 \times 2$, stride 2                             & $8 \times 40 \times 50$   \\
		Conv2D                & $8 \times 40 \times 50$  & $3 \times 3, 16$, stride 1                         & $16 \times 40 \times 50$  \\
		Max Pooling           & $16 \times 40 \times 50$ & $2 \times 2$, stride 2                             & $16 \times 20 \times 25$  \\
		Conv2D                & $16 \times 20 \times 25$ & $3 \times 3, 25$, stride 1                         & $25 \times 20 \times 25$  \\
		Max Pooling           & $25 \times 20 \times 25$ & $2 \times 2$, stride 2                             & $25 \times 10 \times 12$ \\
		Global Avg Pooling    & $25 \times 10 \times 12$ & $10 \times 12$                & $25 \times 1 \times 1$    \\
		Squeeze               & $25 \times 1 \times 1$   & -                                                  & $25$                     \\
		State Feature         & -                        & -                                                  & $8$  \\
		Concatenate           & -                        & -                                                  & $33$ \\
		Fully Connected & 33                       & 128                                                & 128 \\
		Fully Connected & 128                      & 128                                                & 128 \\
		Fully Connected  & 128                   & 4                                                  & 4 \\
		\bottomrule
	\end{tabularx}
\end{table}

This continuous action space design facilitates seamless integration with downstream low-level controllers, such as the differential flatness controller which could calculate the throttle and angular velocity control commands based on the velocity and yaw angle control commands\cite{faessler2017differential}, providing a practical way to implement control commands in real UAV systems.

\subsection{Multi-agent Exploration Strategy}
In this section, we outline the process of multi-agent experience collection. To improve the efficiency of experience gathering and accelerate model convergence, we implement an asynchronous collection strategy. In this setup, multiple environments run in parallel, each with a single UAV interacting with its respective environment, as shown in Figure~\ref{Framework}. These UAVs operate independently, gathering experience through interaction with their respective environments.

By leveraging multiprocessing, we can simultaneously control several UAVs, significantly increasing the rate of experience collection. This enables us to train the model more rapidly, as a larger dataset is generated in a shorter amount of time, leading to faster convergence.

The collected experiences from all UAVs are aggregated to train a central model. This model, in turn, provides decision outputs for each UAV, ensuring consistent policy updates across all environments. By centralizing the learning process, we enable knowledge sharing among the UAVs, which enhances overall performance and improves the success rate of the model. The pseudocode of our proposed DPRL algorithm is shown in Algorithm~\ref{DPRL}.

\begin{algorithm}[H]
	\caption{DPRL: Distributed Privileged Reinforcement Learning}
	\label{DPRL}
	\textbf{Initialize:} Actor network $\pi_\theta$, Critic networks $Q_{\omega_1}, Q_{\omega_2}$, target networks $\pi_{\theta^-}$, $Q_{\omega_1^-}$, $Q_{\omega_2^-}$, experience replay buffer $\mathcal{D}$\\
	\textbf{Hyperparameters:} exploration noise $\epsilon$, batch size $B$, discount factor $\gamma$, target smoothing coefficient $\tau$, target smoothing noise $\xi$
	
	\begin{algorithmic}[1]
		\For{each episode}
		\For{each agent $i$ in $i$th environment}
		\State \textbf{Initialize} state $s_{i,0}$, observation $o_{i,0}$ for Critic network
		\For{each time step $t$}
		\State Select action $a_{i,t} = \pi_\theta(o_{i,t}) + \epsilon$ (with exploration noise)
		\State Execute $a_{i,t}$, get next state $s_{i,t+1}$, next observation $o_{i,t+1}$ and reward $r_{i,t}$
		\State Store $(s_{i,t}, o_{i,t}, a_{i,t}, r_{i,t}, s_{i,t+1}, o_{i,t+1})$ in $\mathcal{D}$
		\EndFor
		\EndFor
		\For{each training step}
		\State Sample mini-batch of $B$ experiences $(s, o, a, r, s', o')$ from $\mathcal{D}$
		\State Compute target action with target policy: $a' = \pi_{\theta^-}(o') + \xi$ (with noise)
		\State Compute target $y = r + \gamma \min_{j=1,2} Q_{\omega_j^-}(s', a')$
		\State Update Critic networks by minimizing loss:
		\[
		L(\omega_j) = \frac{1}{B} \sum_{i=1}^B \left( y - Q_{\omega_j}(s, a) \right)^2, \quad j = 1, 2
		\]
		\If{every $d$ steps}
		\State Update Actor network by maximizing $Q_{\omega_1}(s, \pi_\theta(o))$
		\State Soft update target networks:
		\[
		\theta^- \leftarrow \tau \theta + (1 - \tau) \theta^-, \quad \omega_j^- \leftarrow \tau \omega_j + (1 - \tau) \omega_j^-, \quad j = 1, 2
		\]
		\EndIf
		\EndFor
		\EndFor
	\end{algorithmic}
\end{algorithm}

\section{Experiments and Results}
In this section, extensive experiments were conducted to verify our proposed DPRL navigation algorithm's advantages in terms of efficiency, success rate, and robustness. Our proposed algorithm was compared against TD3 algorithm and EGO-Planner-v2 framework across various environments. Additionally, we performed several ablation studies to validate the novelty of our approach and the rationale behind our choices of state and action spaces. 

\begin{figure}[H]
	\centering
	\begin{minipage}[b]{0.31\linewidth}
		\centering
		\includegraphics[width=\linewidth]{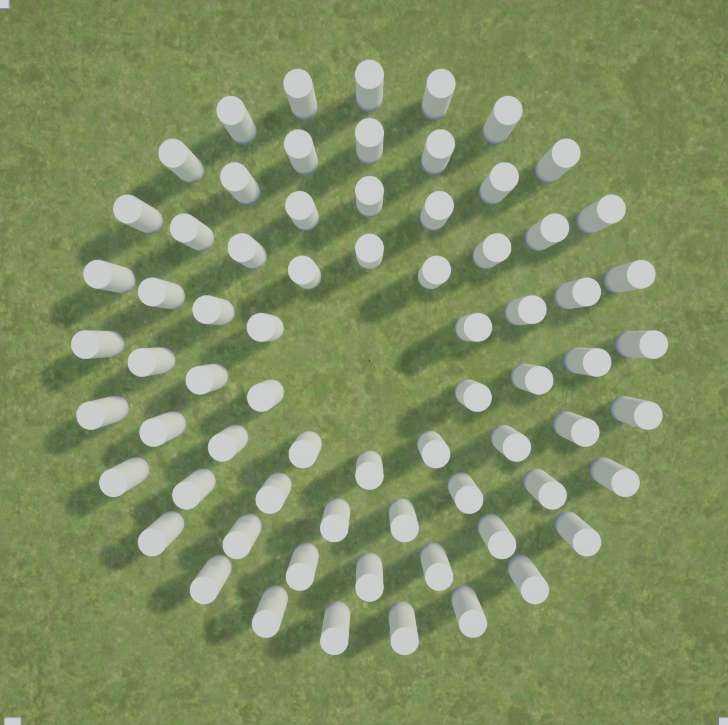}
		\caption*{(\textbf{a})}
	\end{minipage}
	\hfill
	\begin{minipage}[b]{0.31\linewidth}
		\centering
		\includegraphics[width=\linewidth]{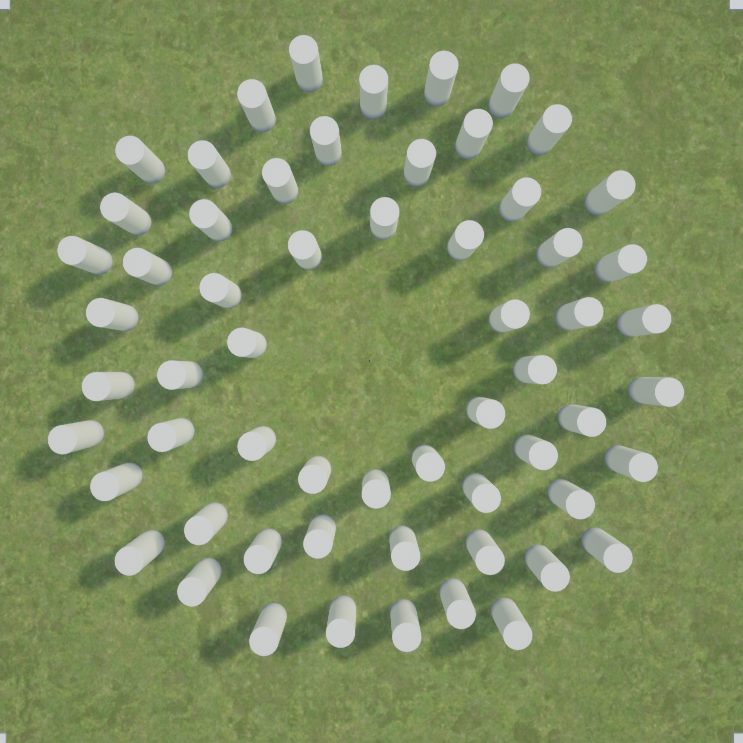}
		\caption*{(\textbf{b})}
	\end{minipage}
	\hfill
	\begin{minipage}[b]{0.31\linewidth}
		\centering
		\includegraphics[width=\linewidth]{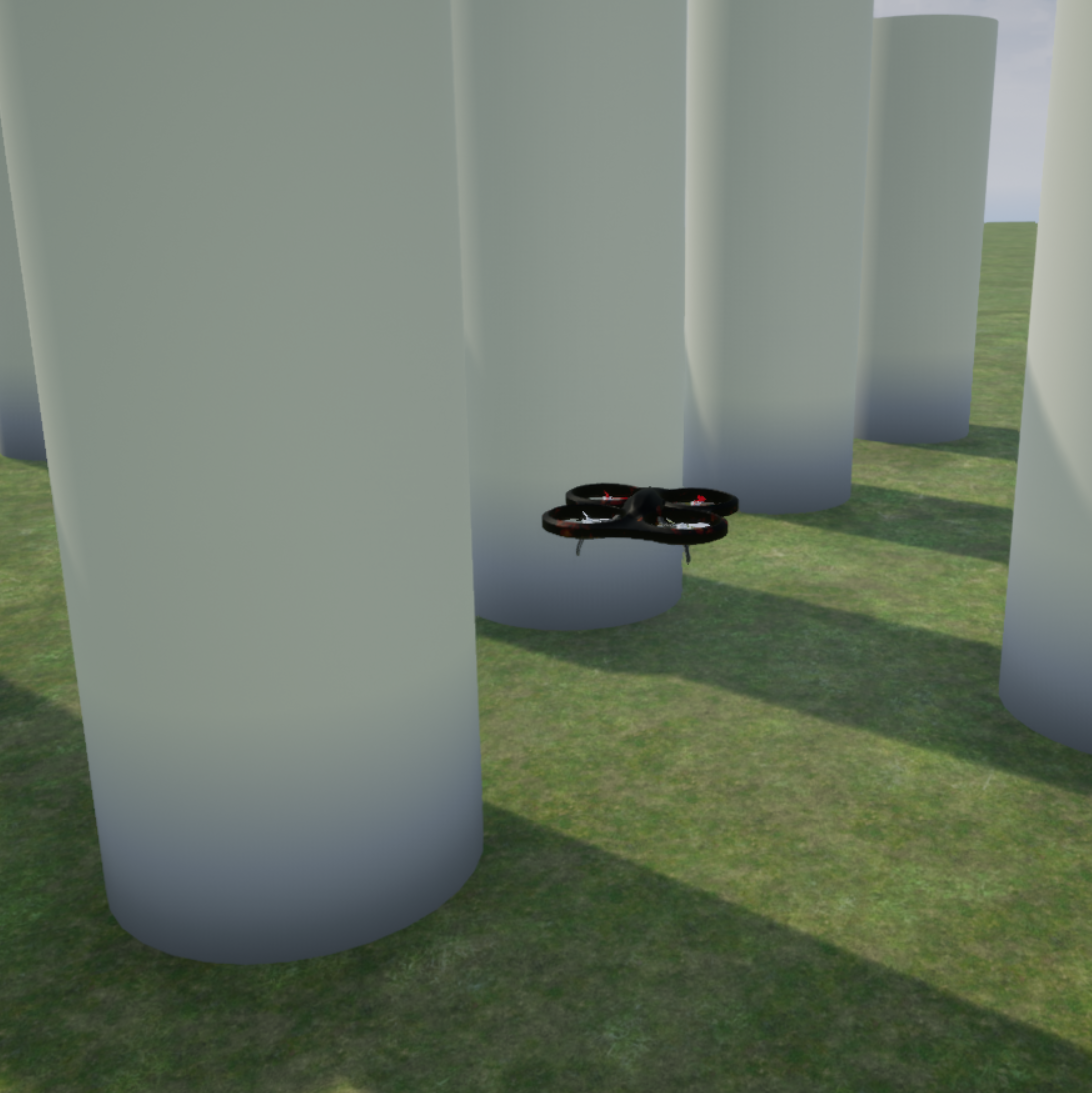}
		\caption*{(\textbf{c})}
	\end{minipage}
	\caption{Simulation environment built using UE4 and AirSim. (\textbf{a}) Top-down view of the training environment. (\textbf{b}) Top-down view of a randomly generated environment. (\textbf{c}) View of the UAV flying within the environment. \label{env}}
\end{figure}

\subsection{Experiments Setup}
We created a realistic simulation environment in UE4, using AirSim's underlying dynamics model for accurate simulation\cite{shah2018airsim}. To emulate a complex obstacle-laden environment, we generated the scene shown in Figure~\ref{env}(a) for model training. This environment features 70 cylindrical obstacles with a radius of 2.5 m and a height of 15 m, arranged within a circular area of 60 m radius centered at the origin. The UAV’s flight altitude is limited to a maximum of 15 m, ensuring it must maneuver around obstacles for collision avoidance rather than flying over them, which would expend excessive energy.

\begin{table}[htbp]
	\caption{Simulation Parameter Settings \label{env_dynamics}}
	\newcolumntype{C}{>{\raggedright\arraybackslash}X}
	\newcolumntype{S}{>{\raggedright\arraybackslash}p{2.5cm}} 
	\begin{tabularx}{\textwidth}{SCC}
		\toprule
		\textbf{Category} & \textbf{Parameter} & \textbf{Value} \\
		\midrule
		\multirow{8}{*}{Environment} 
		& x range & $[-85, 85]$ m \\
		& y range & $[-85, 85]$ m \\
		& z range & $[0.2, 15]$ m \\
		& Start position & $[0,0,5]$ m \\
		& Goal distance & 65 m \\
		& Goal height & 5 m \\
		& Safe distance & 4 m \\
		& Crash distance & 1 m \\
		& Accept radius & 2 m \\
		\midrule
		\multirow{5}{*}{Dynamics} 
		& Action execution duration & 0.1 s \\
		& x-axis velocity range & $[-3.0, 3.0] $ m/s \\
		& y-axis velocity range & $[-3.0, 3.0] $ m/s \\
		& z-axis velocity range & $[-2.0, 2.0] $ m/s \\
		& yaw velocity range & $[-0.3, 0.3]$ rad/s \\
		\midrule
		\multirow{11}{*}{Training} 
		& Discount factor $\gamma$ & 0.99 \\
		& Learning rate $\alpha$ & 3e-4 \\
		& Learning start & 2000 \\
		& Buffer size & 50000 \\
		& Batch size & 128 \\
		& Train frequency & 1 \\
		& Standard deviation of action noise $\sigma_a$ & 0.1 \\
		& Number of environments & 3 \\
		& Total timesteps & 330000 \\
		& Max episode steps & 500 \\
		\midrule
		\multirow{3}{*}{Reward} 
		& $\eta_r$ & 5.0 \\
		& $\eta_p$ & 0.5 \\
		& $\eta_o$ & 1.0 \\
		\midrule
		\multirow{5}{*}{Noise} 
		& $\mu_g$ & 0 \\
		& $\sigma_g$ & 3 \\
		& $p_{sp}$ & 0.005 \\
		& $\mu_s$ & 0 \\
		& $\sigma_s$ & 0.016 \\
		\bottomrule
	\end{tabularx}
\end{table}

Each training episode begins with the UAV taking off from the origin at an initial altitude of 5 meters, aiming for a target position randomly placed along a circumference with a radius of 65 meters. An episode is considered successful when the UAV reaches within 2 m of the target. Conversely, if the UAV comes within 1 m of an obstacle or exits the defined flight area, the episode is marked as a failure.

For each action, an execution duration of 0.1 s is set, ensuring smooth command continuity while avoiding excessive computational load on the simulation, thereby maintaining optimal training frame rates. The specific environmental configurations and dynamics model parameters are outlined in Table~\ref{env_dynamics}.

In our training process, we created three separate environments based on three different seeds to train the model and used an additional environment for model evaluation. UAVs in each environment collected experience independently, controlled by different processes. Training was conducted on a workstation equipped with an Intel i7-13700KF CPU and an NVIDIA 4070 Ti GPU, achieving an average model update frame rate of 20 frames per second. The specific parameters for training, reward function, and noise settings are provided in Table ~\ref{env_dynamics}.

To evaluate the model, we used three metrics: Average Episode Reward (AER), Average Steps of Successful Episodes (ASSE), and Success Rate (SR). AER assesses the overall performance of the algorithm, including navigation accuracy, obstacle avoidance safety, and efficiency. A higher AER reflects better overall performance. ASSE measures the efficiency of the algorithm, where a smaller ASSE value indicates that the UAV completes tasks with fewer steps, demonstrating greater efficiency. SR evaluates the success rate of the algorithm, with a higher SR signifying enhanced safety and practicality.

\subsection{Comparison Experiment}
To validate the comprehensive performance of our UAV autonomous navigation framework, we conducted a comparative analysis of the proposed DPRL algorithm, TD3 algorithm, and EGO-Planner-v2 framework. For DPRL and TD3, we provided noisy visual perception, localization, and other self-state information. In contrast, EGO-Planner received LiDAR point cloud data along with noisy odometry information, with the noise characterized by a mean of 0 and a standard deviation of 0.016. This setup introduced controlled disturbances in the mapping process to simulate real-world conditions. The maximum speed for EGO-Planner was set to be consistent with the RL-based methods, capped at 3 m/s. A PD controller was used for EGO-Planner, which outputs flight commands for velocity and yaw angle. The specific experiments are detailed below.

\begin{figure}[H]
	\centering
	\begin{minipage}[b]{0.47\linewidth}
		\centering
		\includegraphics[width=\linewidth]{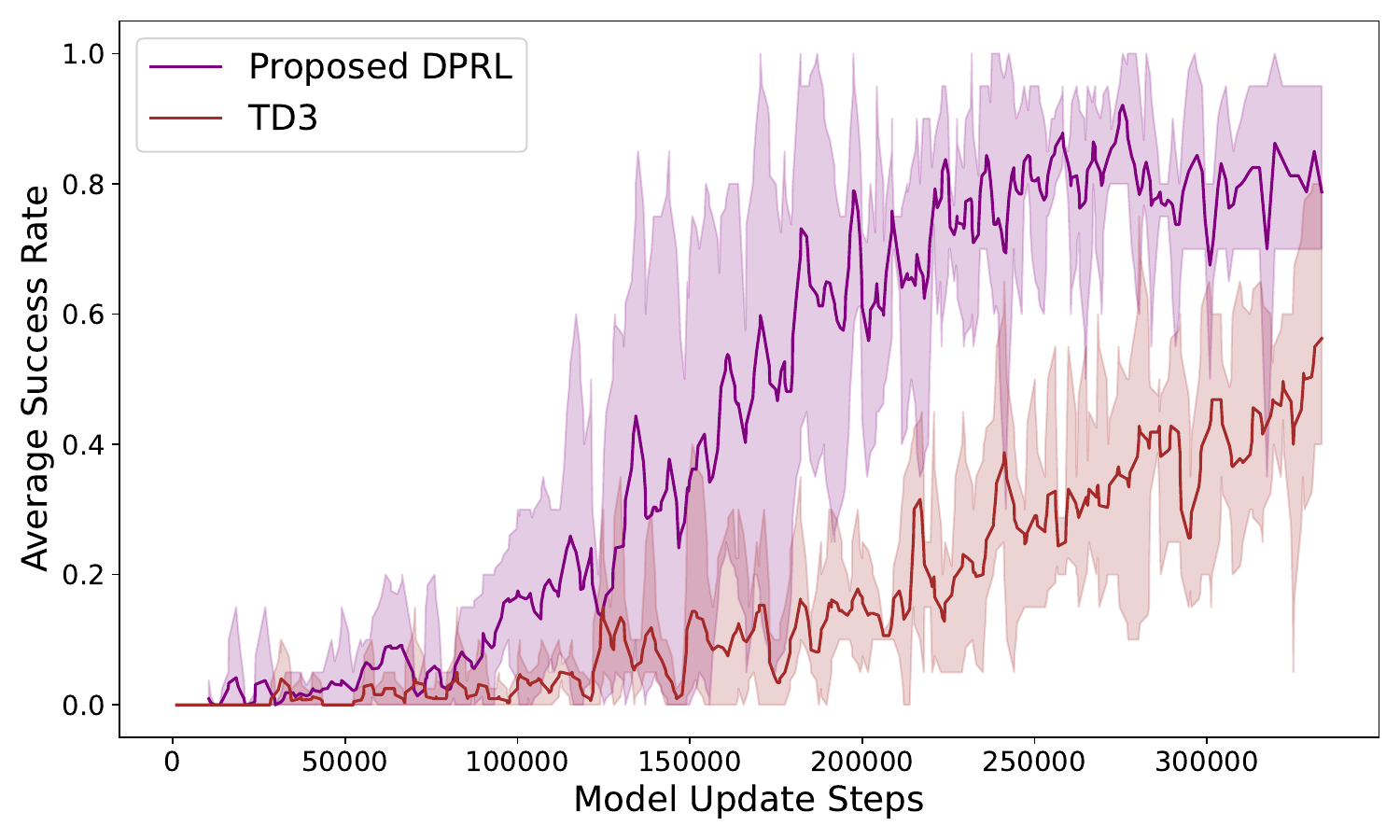}
		\caption*{(\textbf{a})}
	\end{minipage}
	\hfill
	\begin{minipage}[b]{0.47\linewidth}
		\centering
		\includegraphics[width=\linewidth]{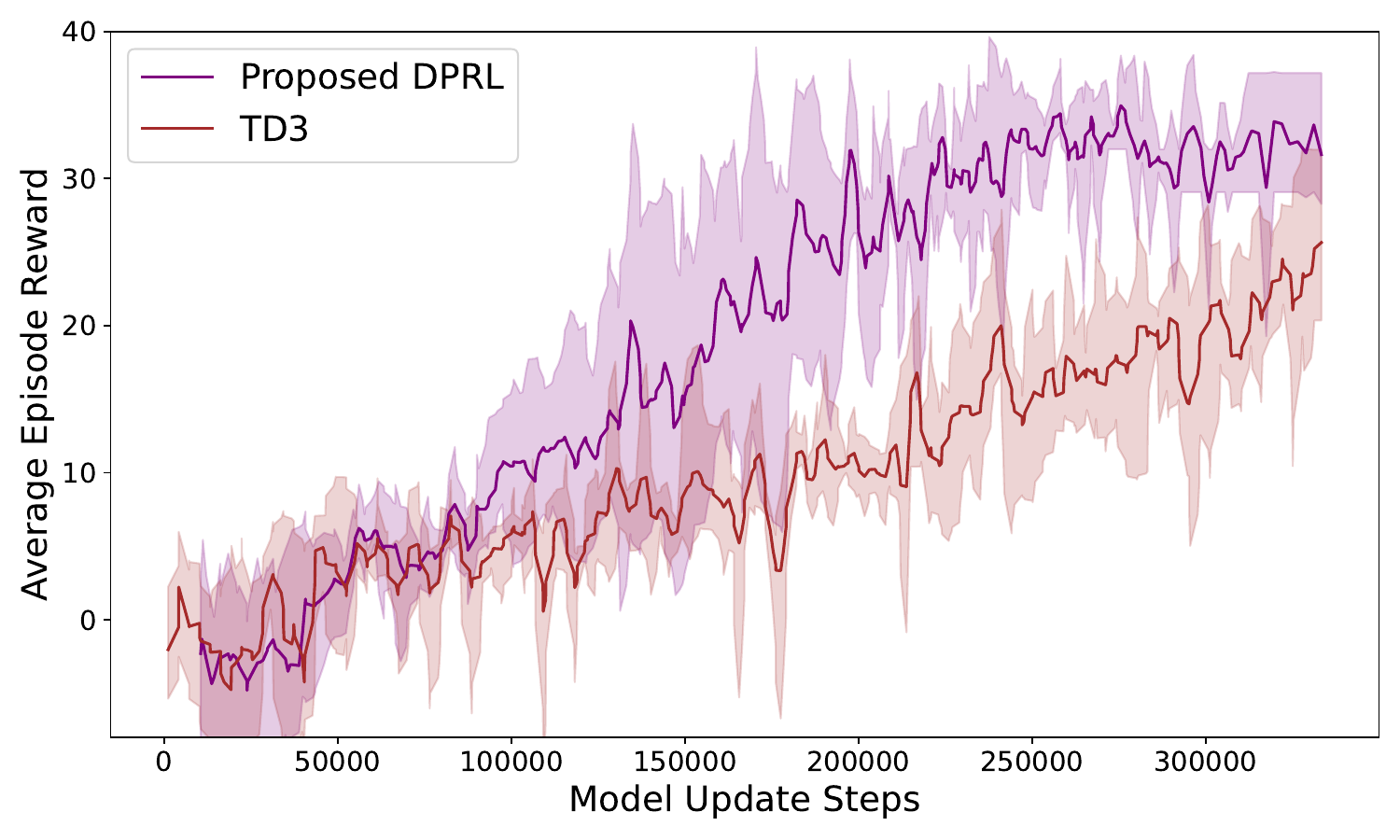}
		\caption*{(\textbf{d})}
	\end{minipage}
	\caption{The training curves of different UAV navigation algorithms. (\textbf{a}) Average success rate curve of Proposed DPRL and TD3. (\textbf{b}) Average  episode reward curve of Proposed DPRL and TD3. \label{res1_1}}
\end{figure}

Figure~\ref{res1_1}(a) and (b) present the comparative training curves between DPRL and TD3 under four different seeds. As illustrated in Figure~\ref{res1_1}(a), DPRL demonstrates a notably faster convergence rate compared to TD3, reaching an average success rate exceeding 85\% after just 220,000 training steps. The average episode reward curves during training, shown in Figure~\ref{res1_1}(b), align closely with the trends observed in the average success rate curves. DPRL exhibits rapid reward growth during the early and mid-stages of training, stabilizing at a high reward value after 240,000 steps. In contrast, TD3 shows consistently slower reward growth and fails to converge by the end of training.

\begin{figure}[t]
	\centering
	\begin{minipage}[b]{0.31\linewidth}
		\centering
		\includegraphics[width=\linewidth]{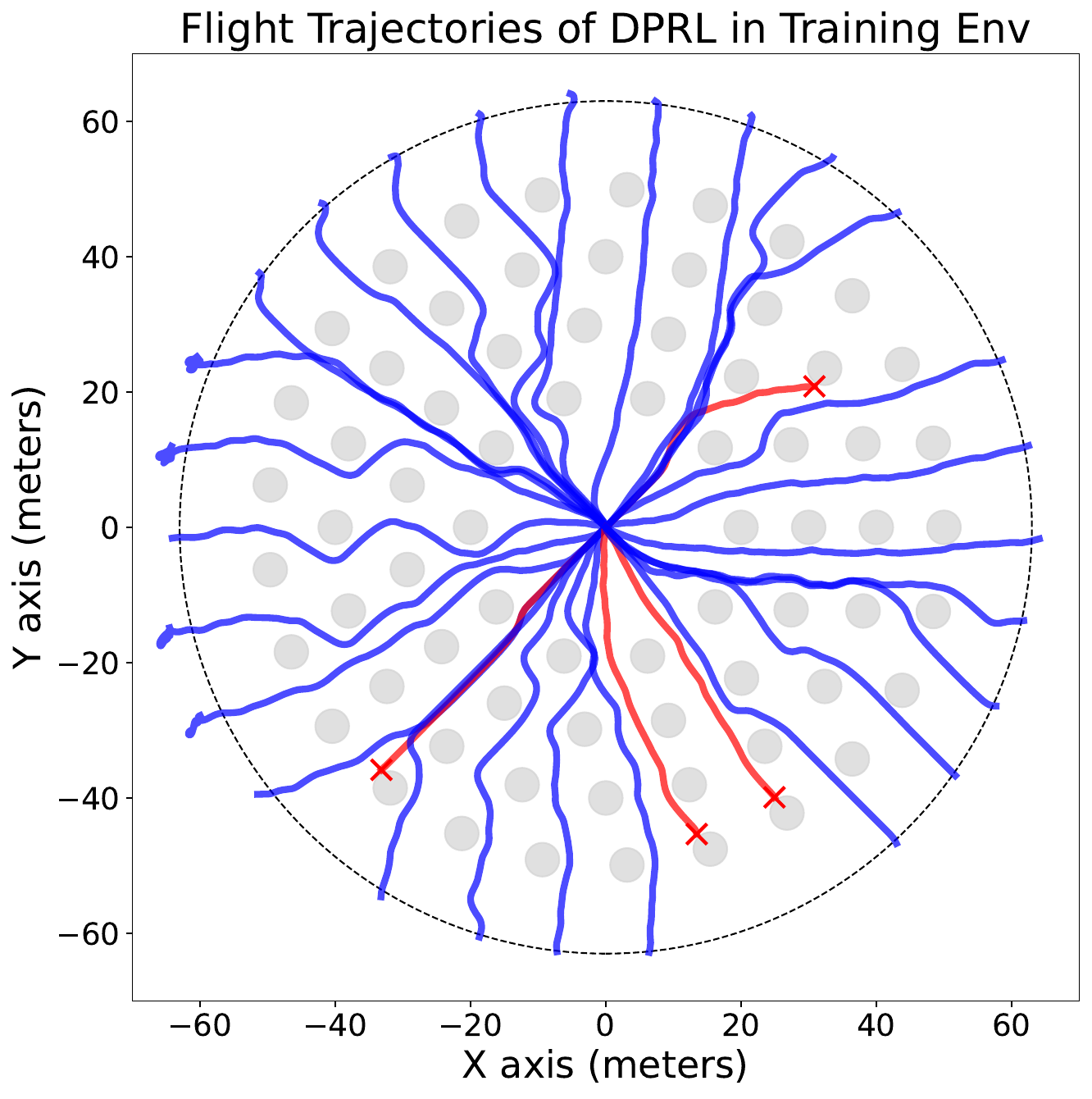}
		\caption*{(\textbf{a})}
	\end{minipage}
	\hfill
	\begin{minipage}[b]{0.31\linewidth}
		\centering
		\includegraphics[width=\linewidth]{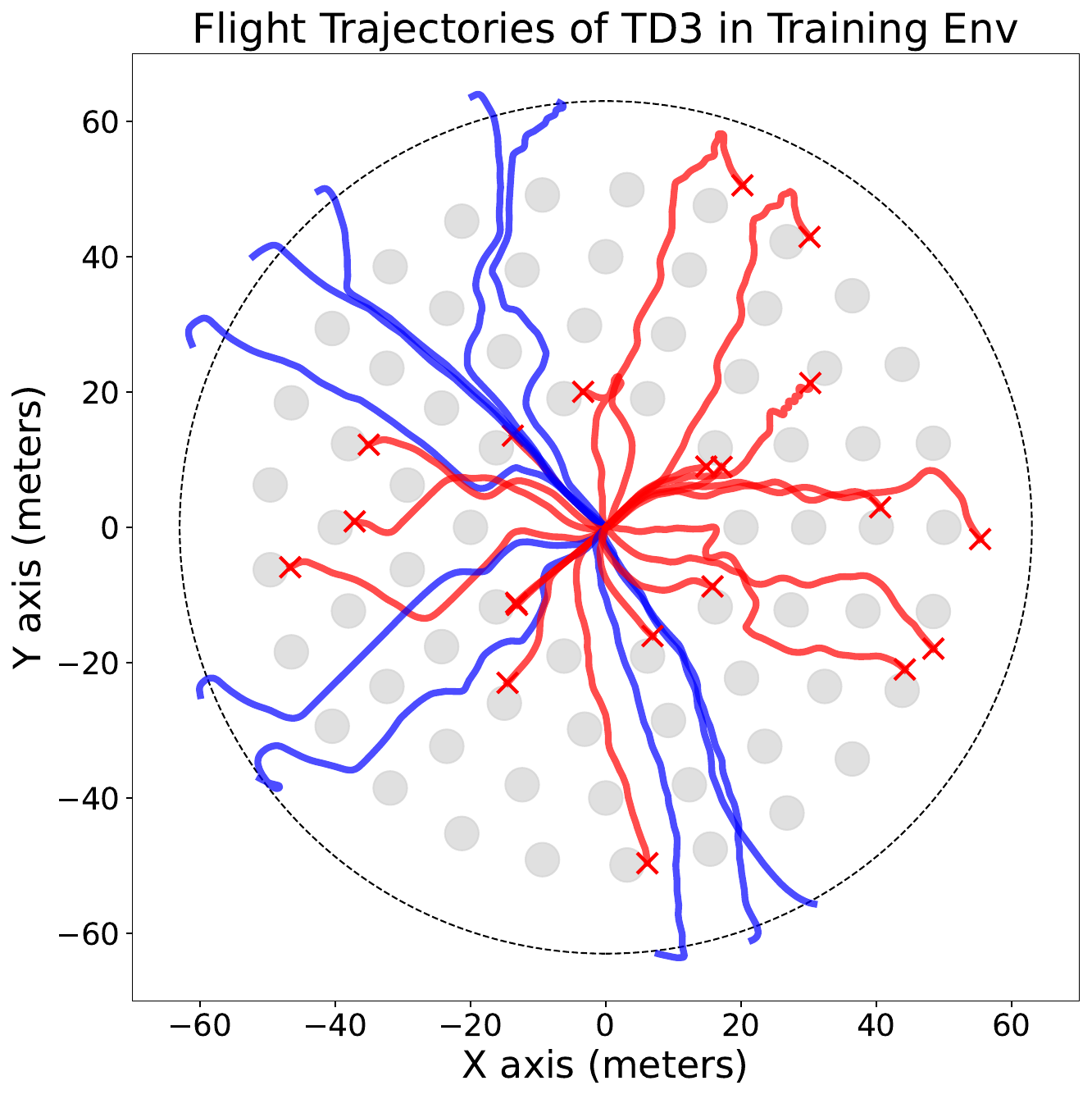}
		\caption*{(\textbf{b})}
	\end{minipage}
	\hfill
	\begin{minipage}[b]{0.31\linewidth}
		\centering
		\includegraphics[width=\linewidth]{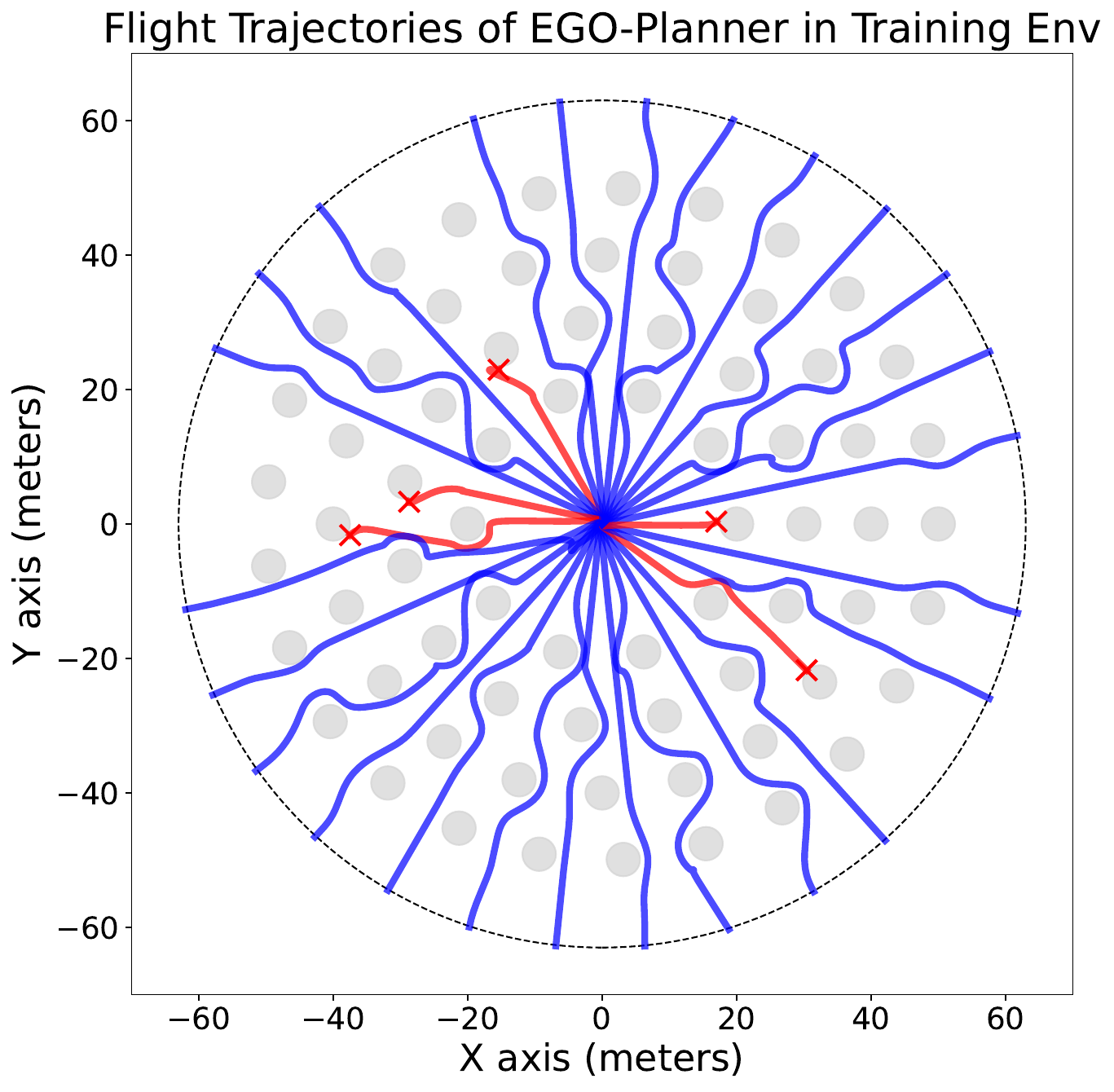}
		\caption*{(\textbf{c})}
	\end{minipage}
	\vskip 0.4cm
	\begin{minipage}[b]{0.31\linewidth}
		\centering
		\includegraphics[width=\linewidth]{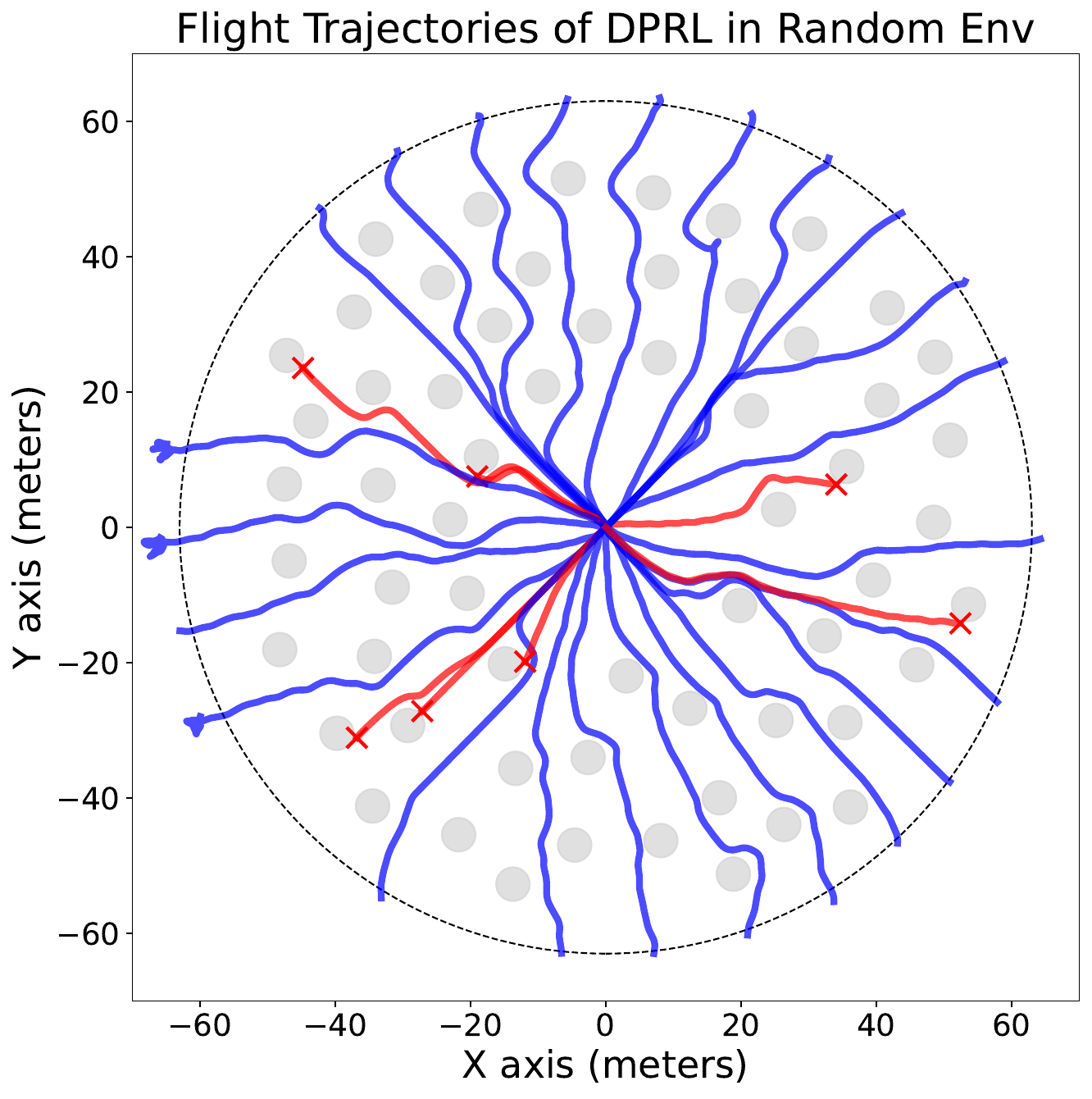}
		\caption*{(\textbf{d})}
	\end{minipage}
	\hfill
	\begin{minipage}[b]{0.31\linewidth}
		\centering
		\includegraphics[width=\linewidth]{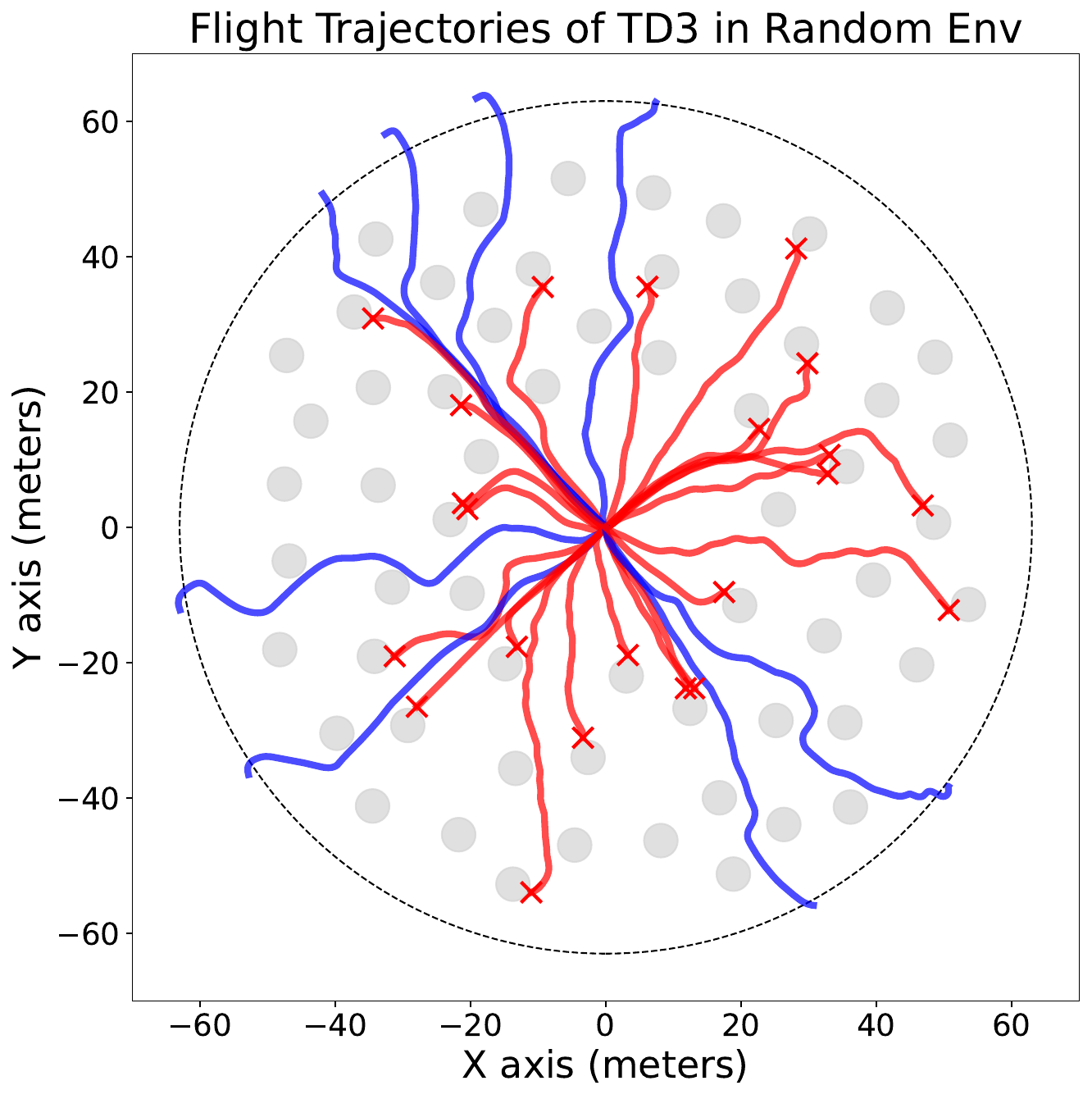}
		\caption*{(\textbf{e})}
	\end{minipage}
	\hfill
	\begin{minipage}[b]{0.31\linewidth}
		\centering
		\includegraphics[width=\linewidth]{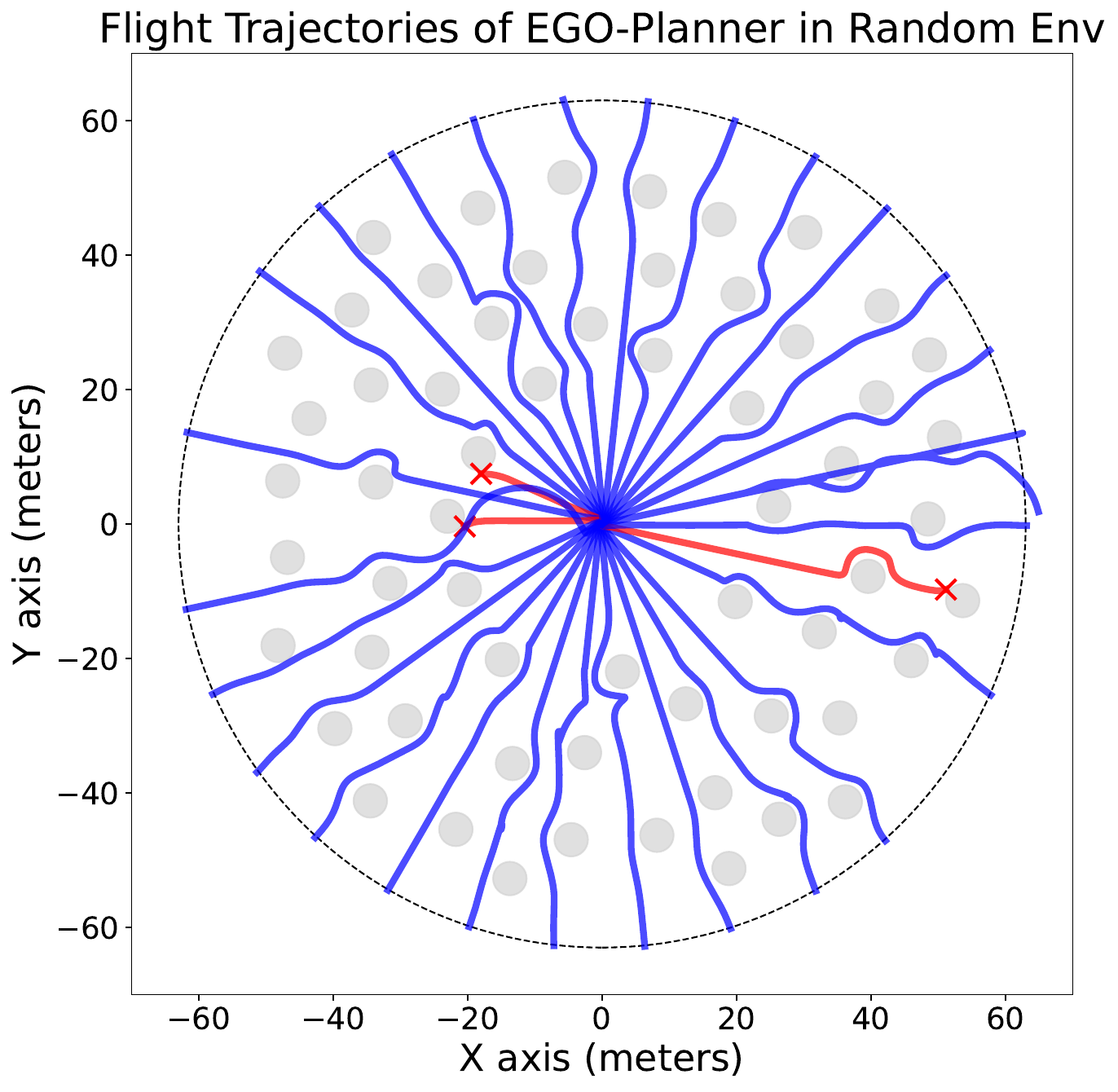}
		\caption*{(\textbf{f})}
	\end{minipage}
	\caption{Comparison results of navigation and obstacle avoidance trajectories. Each figure represents the evaluation results of 30 episodes, where blue trajectories indicate successful completions, and red trajectories represent failures due to collisions. (\textbf{a}) Flight trajectories of DPRL in training environment. (\textbf{b}) Flight trajectories of TD3 in training environment. (\textbf{c}) Flight trajectories of EGO-Planner-v2 in training environment. (\textbf{d}) Flight trajectories of DPRL in random environment. (\textbf{e}) Flight trajectories of TD3 in random environment. (\textbf{f}) Flight trajectories of EGO-Planner-v2 in random environment. \label{res2}}
\end{figure}

We deployed the trained models of DPRL, TD3, and EGO-Planner-v2 in both the training and randomly generated environments(Figure~\ref{env}(b)), plotting the flight trajectories over 30 episodes with unique target positions, as shown in Figure~\ref{res2}. It can be observed that DPRL maintains a high success rate in both the training and random environments. Although the success rate decreases somewhat in unfamiliar environments compared to the training environment, the model still demonstrates robustness against noise interference, underscoring its adaptability to new environments.

In contrast, TD3 exhibits the lowest success rate, performing poorly in both environments, with less smooth trajectories compared to DPRL. For EGO-Planner, both environments are unfamiliar, with the training environment containing a higher density of obstacles. Consequently, EGO-Planner’s performance is weaker in the training environment than in the random environment. EGO-Planner produces the smoothest trajectories and reaches target points with the highest accuracy, but its planning speed is the slowest among the three algorithms.

\begin{table}[H]
	\caption{Performance Comparison in Training and Random Environments \label{env_performance}}
	\centering
	\newcolumntype{C}{>{\raggedright\arraybackslash}X}
	\newcolumntype{S}{>{\raggedright\arraybackslash}p{2.8cm}} 
	\newcolumntype{M}{>{\raggedright\arraybackslash}p{4.5cm}} 
	\begin{tabularx}{\textwidth}{SScccc}
		\toprule
		\textbf{Environment} & \textbf{Algorithm} & \multicolumn{3}{c}{\textbf{Metric}} \\ 
		\cmidrule(l){3-5}
		&  & \textbf{AER} & \textbf{ASSE} & \textbf{SR} \\ 
		\midrule
		\multirow{3}{*}{Training Env} 
		& DPRL & \textbf{33.685} & \textbf{191.0} & \textbf{86.667\%} \\ 
		& TD3 & 16.205 & 202.0 & 33.333\% \\ 
		& EGO-Planner-v2 & 22.746 & 399.625 & 83.333\% \\ 
		\midrule
		\multirow{3}{*}{Random Env} 
		& DPRL & \textbf{31.617} & \textbf{190.391} & 76.667\% \\ 
		& TD3 & 17.033 & 196.875 & 26.667\% \\ 
		& EGO-Planner-v2 & 24.527 & 374.0 & \textbf{90.0\%} \\ 
		\bottomrule
	\end{tabularx}
\end{table}

Table~\ref{env_performance} summarizes the average episode reward, average steps of successful episodes, and success rate of each algorithm. While DRPL achieves a success rate comparable to EGO-Planner, it attains higher rewards and shorter episode lengths, demonstrating superior obstacle-avoidance efficiency and best overall performance among the three methods.

\subsection{Ablation Experiment}
\begin{figure}[t]
	\begin{minipage}[b]{0.47\linewidth}
		\centering
		\includegraphics[width=\linewidth]{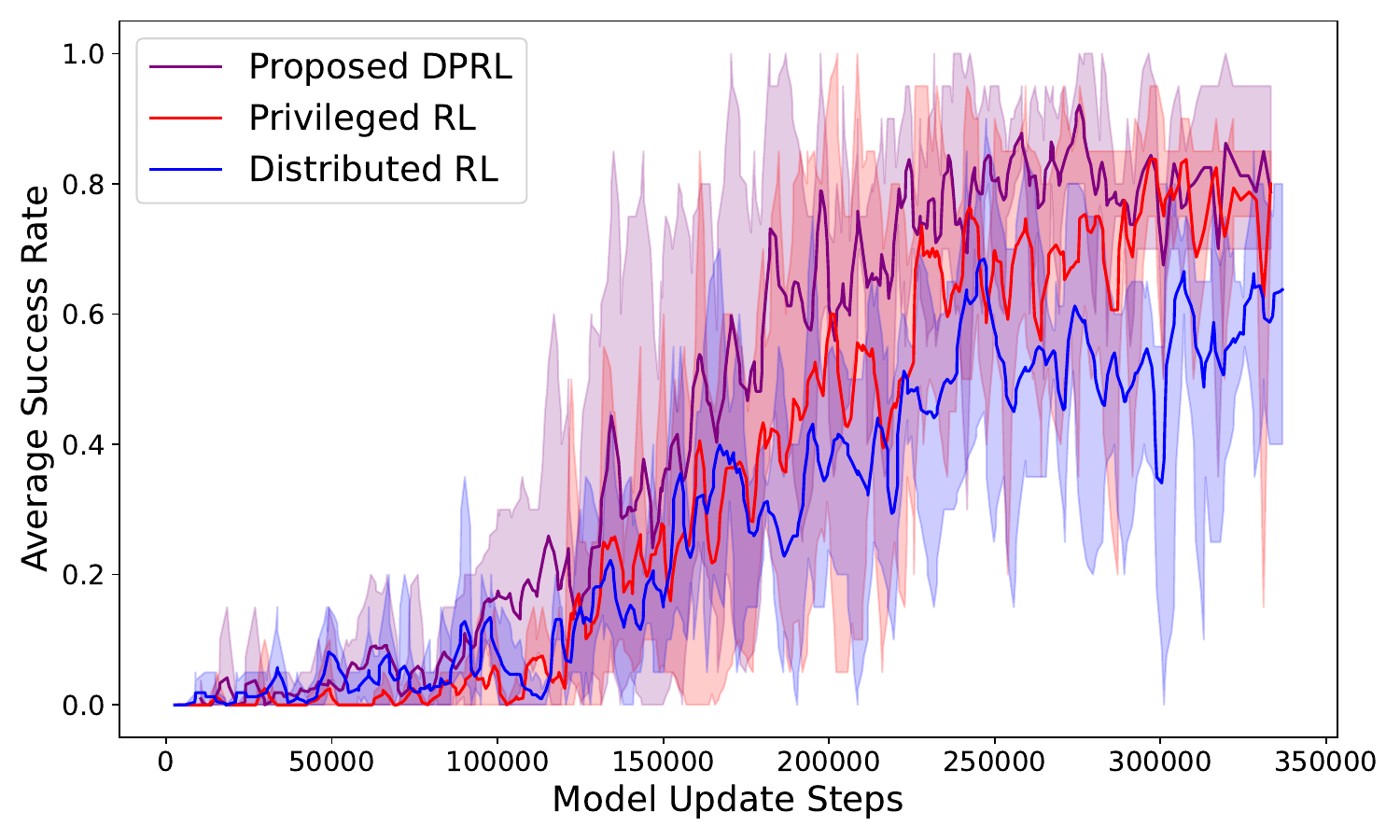}
		\caption*{(\textbf{a})}
	\end{minipage}
	\hfill
	\begin{minipage}[b]{0.47\linewidth}
		\centering
		\includegraphics[width=\linewidth]{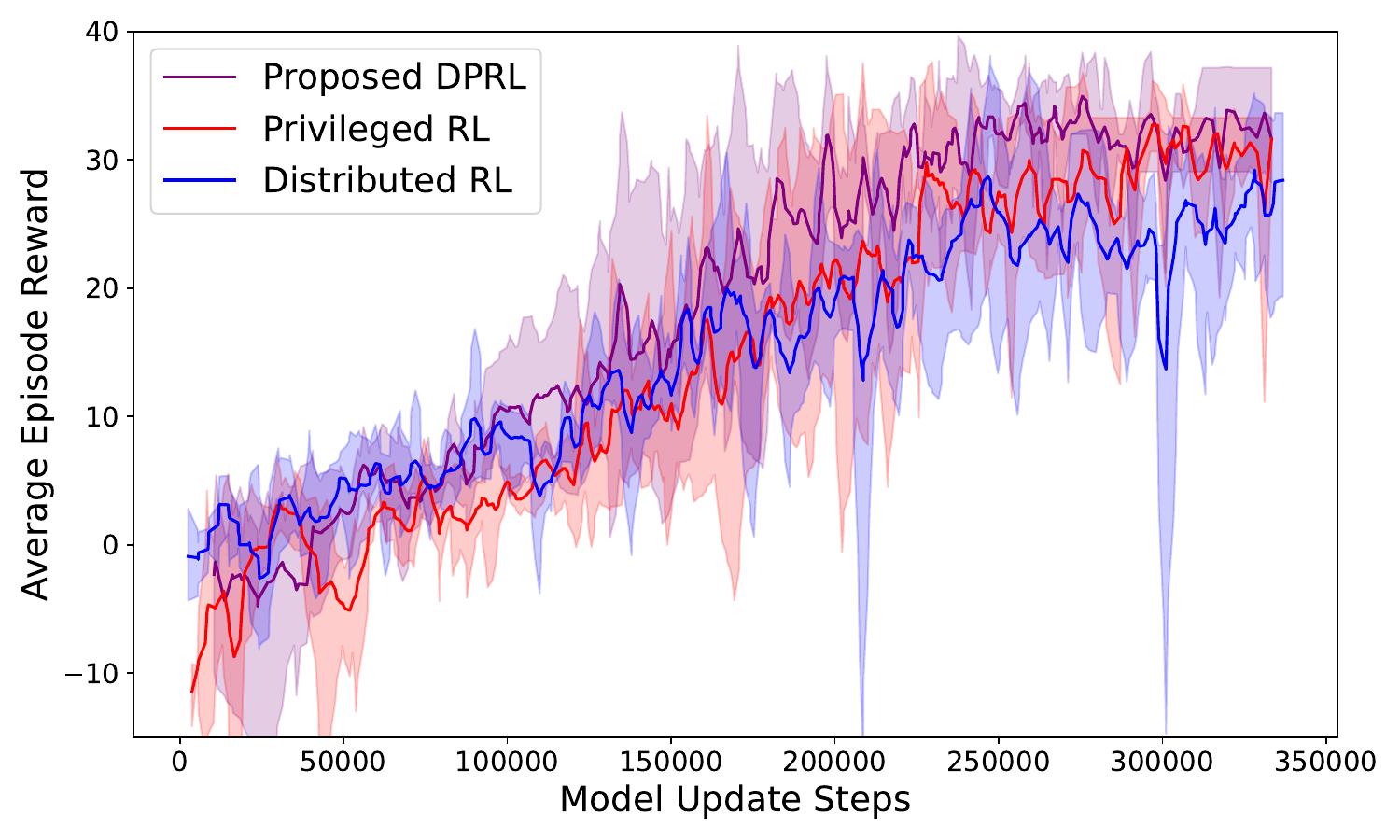}
		\caption*{(\textbf{b})}
	\end{minipage}
	\caption{Ablation experiment results for key components in DPRL during model training. (\textbf{a}) Average success rate curve of Proposed DPRL, Privileged RL and Distributed RL. (\textbf{b}) Average episode reward curve of Proposed DPRL, Privileged RL and Distributed RL. \label{res1_2}}
\end{figure}

We first conducted ablation studies on the privileged learning and multi-agent exploration components of the proposed DPRL algorithm to validate their effects. The detailed experimental results are shown in Figures~\ref{res1_2}(a) and (b). In these studies, Privileged RL and Distributed RL were derived from DPRL by removing multi-agent exploration and privileged learning, respectively.

As shown in Figures~\ref{res1_2}(a), DPRL demonstrates superior convergence speed and a higher final average success rate during training compared to both Privileged RL and Distributed RL. Specifically, DPRL's success rate stabilizes after 220,000 steps, while both Privileged RL and Distributed RL only converge after 300,000 steps. The average episode reward curves, shown in Figures~\ref{res1_2}(b), align closely with the average success rate trends. DPRL's rewards stabilize above 30 after 240,000 steps, whereas Privileged RL requires over 300,000 steps to achieve similar rewards. In contrast, Distributed RL fails to exceed a reward of 30 even by the end of training. 

Notably, the impact of privileged learning is greater than that of multi-agent exploration. Although Distributed RL shows slightly better convergence than Privileged RL at the beginning of training, Privileged RL begins to outperform after 200,000 training steps, ultimately achieving a higher average success rate and average episode reward. This highlights the significant advantage of privileged learning in handling partial observability in the environment and the effectiveness of multi-agent exploration in accelerating early convergence.

To compare the impact of different state and action space designs on the model, we then conducted the following experiment. We established an alternative state and action space similar to that used by He et al.\cite{he2020deep}, with modifications to the self-state vector's positioning and velocity information. Specifically, the state now includes the distance to the target in the xy-plane, the z-axis distance, the velocity in the xy-plane, and the z-axis velocity. This adjustment reduces the total dimensionality of the state vector from 33 to 31. Correspondingly, the action space was reduced from 4 dimensions to 3. The original x and y velocity components were replaced by a single xy-plane velocity, which is split into x and y components based on the current yaw angle during execution. This setup ensures that the UAV always flies in the direction it is facing, keeping obstacles in its forward field of view. However, this also significantly compresses the UAV's action space, reducing maneuverability.

\begin{figure}[t]
	\begin{minipage}[b]{0.47\linewidth}
		\centering
		\includegraphics[width=\linewidth]{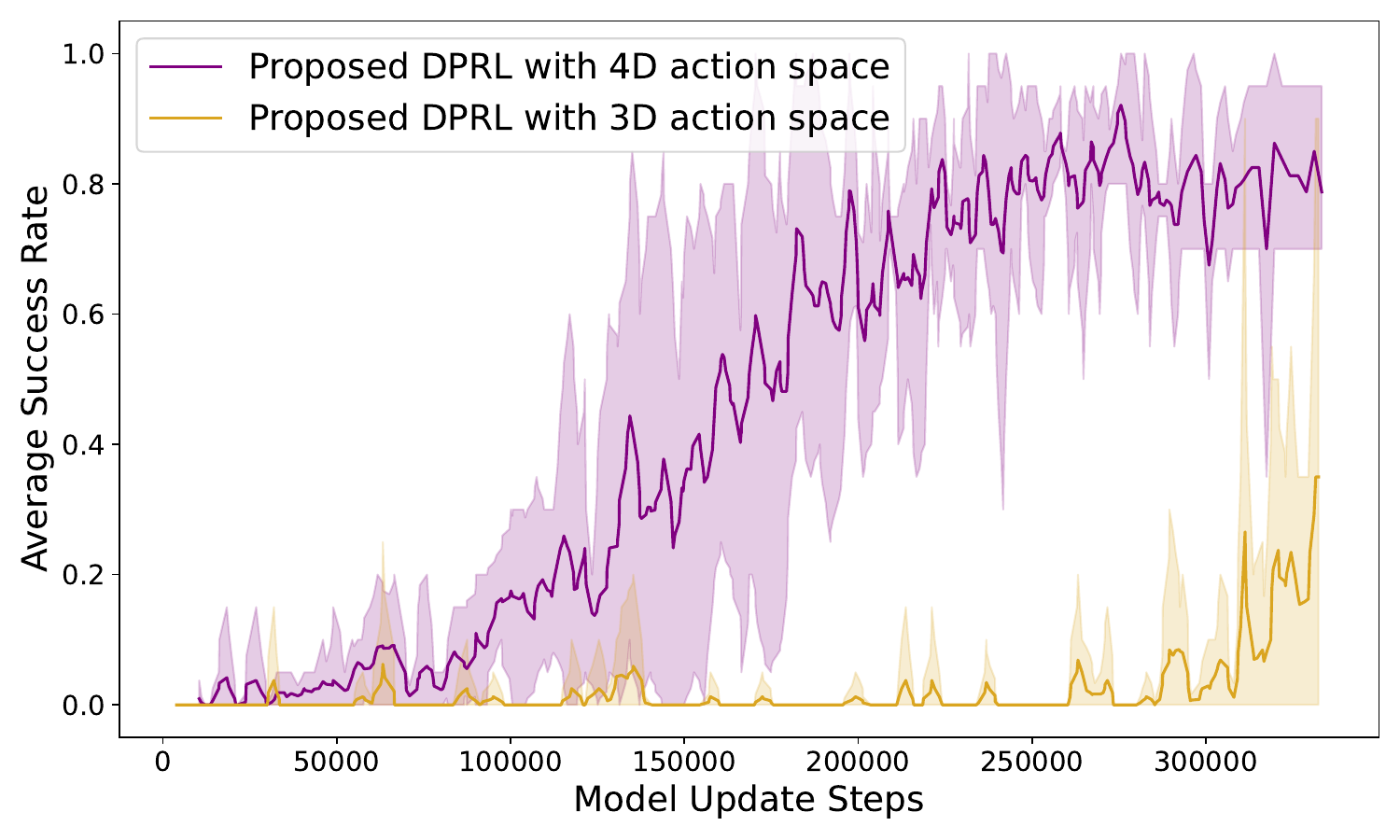}
		\caption*{(\textbf{a})}
	\end{minipage}
	\hfill
	\begin{minipage}[b]{0.47\linewidth}
		\centering
		\includegraphics[width=\linewidth]{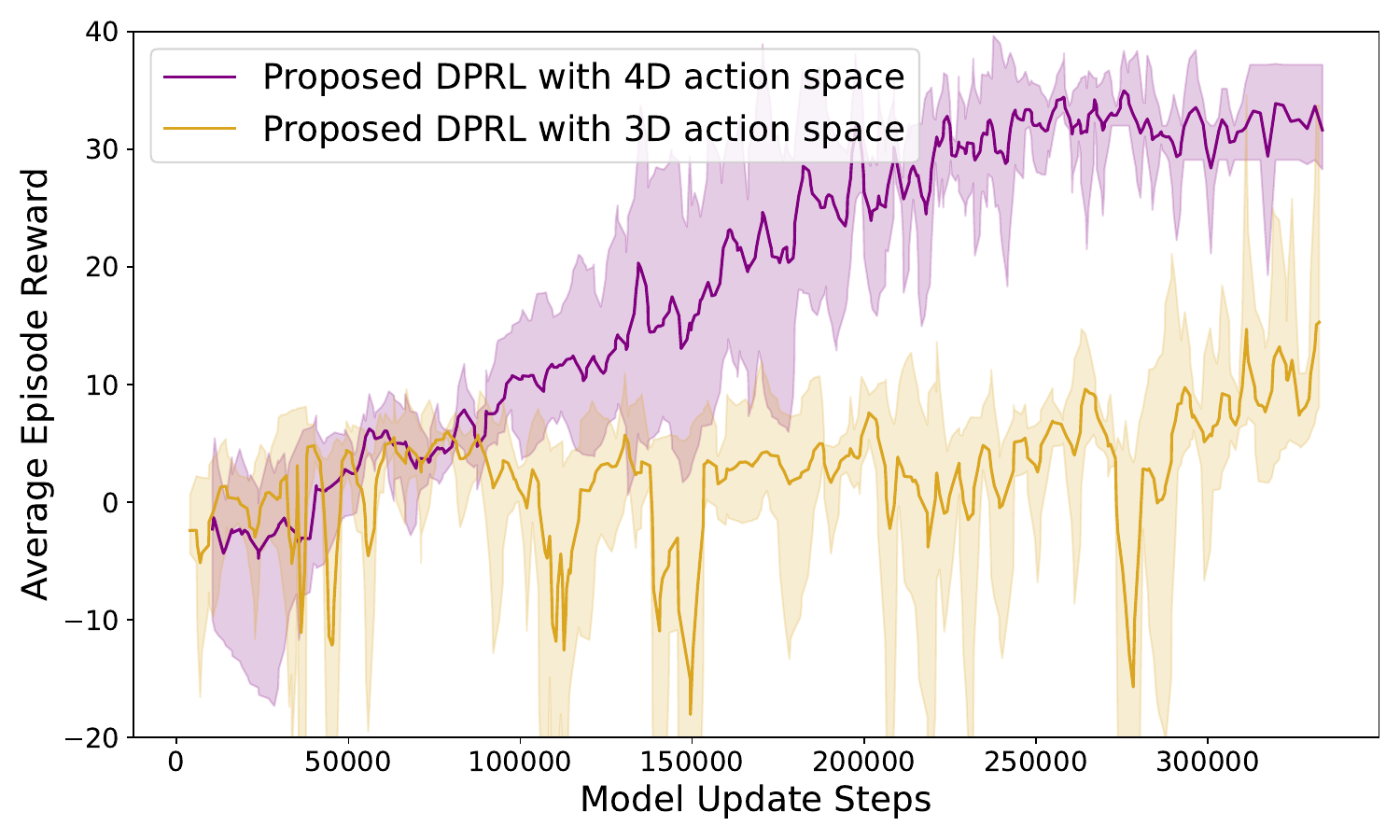}
		\caption*{(\textbf{b})}
	\end{minipage}
	\caption{Ablation experiment results for state and action space design in DPRL during model training. (\textbf{a}) Average success rate curve of DPRL with 3D and 4D action space. (\textbf{b}) Average episode reward curve of DPRL with 3D and 4D action space.\label{res_final}}
\end{figure}

The actual training and evaluation results are shown in Figures~\ref{res_final}. It is evident that our proposed 4-dimensional action space, along with its corresponding state space, significantly outperforms the 3-dimensional action space and its associated state space. During training, the DPRL model with the 3-dimensional action space exhibited almost no successful flight episodes in the early stages and demonstrated very slow learning progress. By the end of training, it achieved only an average success rate of about 30\%, performing even worse than TD3 with the 4-dimensional action space. This poor performance is further reflected in the average episode reward results. The DPRL model with the 3-dimensional action space showed minimal improvement in rewards throughout training and only reached an average reward of 10 by the end, highlighting the substantial negative impact of action space compression on the model’s ability to learn obstacle-avoidance navigation. These findings confirm the rationality and effectiveness of our proposed state and action space design.

\section{Conclusions}
In our work, we propose the DPRL algorithm for UAV autonomous navigation and obstacle avoidance in complex, unknown environments. Specifically, we implement privileged learning using an asymmetric Actor-Critic network structure to address perception and localization noise encountered in flight environments. Additionally, we utilize asynchronous multi-agent exploration across multiple environments to improve data efficiency and accelerate model convergence.

Experiments conducted in the AirSim simulation environment demonstrate the DPRL algorithm's comprehensive optimal performance in terms of convergence speed, final flight success rate, robustness to environmental variations, and planning efficiency. The results also validate the effectiveness of our novelty and the designed state and action space. Moreover, our algorithm shows strong potential for transfer to real-world applications and is compatible with all off-policy deep reinforcement learning algorithms, ensuring broad generalizability.

In future work, we will further enhance the navigation and obstacle avoidance success rate of the DPRL algorithm across various complex environments and conduct outdoor flight experiments to validate the proposed algorithm more comprehensively.

\end{document}